\documentclass[sigconf]{acmart}

\usepackage{url}
\usepackage{booktabs}
\usepackage{multirow}
\usepackage{graphicx}
\usepackage{enumitem}
\usepackage[table,xcdraw]{xcolor}
\usepackage{array}
\usepackage{float}

\usepackage[capitalize]{cleveref}
\crefname{section}{Sec.}{Secs.}
\Crefname{section}{Section}{Sections}
\Crefname{figure}{Figure}{Figures}
\crefname{figure}{Fig.}{Figs.}
\Crefname{table}{Table}{Tables}
\crefname{table}{Tab.}{Tabs.}
\Crefname{algorithm}{Algorithm}{Algorithms}
\crefname{algorithm}{Alg.}{Algs.}
\Crefname{Equation}{Equation}{Equations}
\crefname{Equation}{Eq.}{Eqs.}
\Crefname{Appendix}{Appendix}{Appendices}
\crefname{Appendix}{App.}{Apps.}
\Crefname{Assumption}{Assumption}{Assumptions}
\Crefname{Corollary}{Corollary}{Corollary}

\usepackage[caption=false, font=normalsize, labelfont=sf, textfont=sf]{subfig}

\usepackage[usenames,dvipsnames]{xcolor}

\newcommand{\x}{\mathbf{x}}
\renewcommand{\v}{\mathbf{v}}
\newcommand{\net}{\text{\texttt{net}}}

\AtBeginDocument{%
  }

\setcopyright{acmlicensed}
\copyrightyear{2018}
\acmYear{2018}
\acmDOI{XXXXXXX.XXXXXXX}
\acmConference[Conference acronym 'XX]{Make sure to enter the correct
  conference title from your rights confirmation email}{June 03--05,
  2018}{Woodstock, NY}

\begin{document}

\title{More Genes, Faster Inference: Efficient Large-Panel Spatial Transcriptomics Prediction from Histology}
\author{Songming Zhang}
\authornote{Equal contribution.}
\email{sm.zhang@smail.nju.edu.cn}
\affiliation{%
  \institution{School of Computer Science, \\ Nanjing University}
  \institution{Faculty of Computer Engineering and Artificial Intelligence, Shenzhen University of Advanced Technology}
  \city{Nanjing}
  \country{China}
  }

\author{Yuxiao Luo}
\authornotemark[1]
\email{yuxiao.luo@connect.polyu.hk}
\affiliation{%
  \institution{Department of Land Surveying and Geospatial Science (LSGS), The Hong Kong Polytechnic University}
  \city{Hong Kong}
  \country{China}
  }
\author{Shifu Luo}
\email{yc37182@umac.mo}
\affiliation{%
  \institution{Faculty of Health Sciences, \\ University of Macau}
  \institution{Faculty of Computer Engineering and Artificial Intelligence, Shenzhen University of Advanced Technology}
  \city{Macau}
  \country{China}
  }

\author{Siran Chen}
\email{srancchen@tencent.com}
\affiliation{%
  \institution{Tencent}
  \city{Shenzhen}
  \country{China}
}

\author{Shuquan Su}
\email{sushuquan@suat-sz.edu.cn}
\affiliation{%
\institution{Faculty of Computer Engineering and Artificial Intelligence, Shenzhen University of Advanced Technology}
  \city{Shenzhen}
  \country{China}
}

\author{Jinyan Li}
\authornote{Corresponding author: Jinyan Li.}
\email{lijinyan@suat-sz.edu.cn}
\affiliation{%
    \institution{Faculty of Computer Engineering and Artificial Intelligence, Shenzhen University of Advanced Technology}
    \city{Shenzhen}
  \country{China}
}
\renewcommand{\shortauthors}{Zhang et al.}

\begin{abstract}
Spatial transcriptomics (ST) prediction from histology aims to infer spatially resolved gene expression profiles directly from tissue images.
This is a challenging problem that is currently often addressed with complicated model architectures, additional inputs, and complicated training procedures, while remaining limited to fixed, small gene panels.
This raises a natural question:
\emph{Is there an efficient method for large-panel ST prediction?}
Here, we present GeneFlow, a simpler yet effective framework for large-scale spatial gene expression prediction that unifies flow-matching generative modeling with structured multimodal fusion.
Specifically, GeneFlow first encodes both histological morphology and cross-gene interactions in spatial spots through histology and gene tokenizers.
The model directly predicts spatial gene expression, rather than an intermediate velocity field, while being trained with a velocity-based flow-matching objective.
We show that, even without relying on heavy generative pipelines, GeneFlow achieves state-of-the-art performance across multiple ST benchmarks, with up to 56.8\% gains in 1,000-gene prediction.
Extensive experiments further show that GeneFlow produces biologically meaningful spot profiles while reducing computational cost by orders of magnitude, thereby improving efficiency and scalability, particularly in clinical settings.
We believe GeneFlow provides an opportunity for leveraging widely available histology images for downstream biological discovery.
Code is available at: \url{https://github.com/ZZhangsm/GeneFlow}
\end{abstract}


\maketitle

\section{Introduction}
Histopathology with hematoxylin and eosin (H\&E) staining has been the gold standard for clinical diagnosis of cancer and other diseases for more than a century. However, microscopic examination of H\&E images is often time-consuming and cannot reliably distinguish many molecularly distinct cell types by visual inspection alone.
Spatial transcriptomics (ST) technologies have enabled the joint measurement of tissue morphology and gene expression over the past decade~\citep{staahl2016visualization,chen2020spatial}.
For instance, spot-based platforms such as 10x Visium~\citep{janesick2023high} can quantify over 20,000 genes across entire sections without tissue dissociation.
By providing spatially resolved gene expression profiles, ST platforms enable new insights into the prediction of key clinical outcomes of complex cellular behavior and disease states~\citep{baccin2020combined,jiang2026stwave}.

\begin{figure*}[tb]
	\centering
	\includegraphics[width=.95\linewidth]{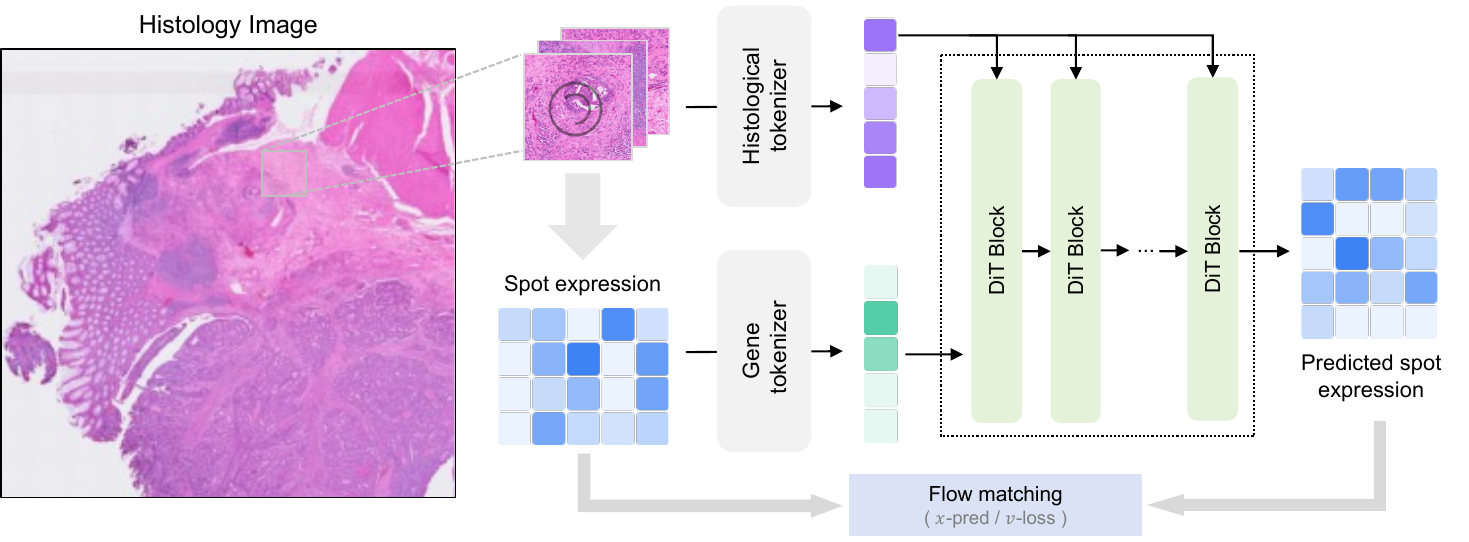}
	\caption{Overview of GeneFlow.
    The training data for GeneFlow consists of ST datasets with paired H\&E histology images and spot-level expression profiles. Gene expressions are first tokenized into latent representations, while image patches centered at each spot are encoded using a pretrained pathology foundation model to provide conditioning features. These image features guide a genetic flow-based network to learn endpoint-based gene expression prediction with velocity loss.
    At inference time, GeneFlow iteratively predicts gene expression from a given histology image patch.}
	\label{fig:overview}
\end{figure*}

However, despite recent advances in ST technologies, significant challenges remain for clinical translation, including high cost and experimental complexity.
In contrast, histopathology images are routinely acquired and widely available in clinical practice.
H\&E-stained tissue sections provide rich structural information that reflects underlying biological processes at both cellular and tissue levels.
Motivated by this complementarity between ST and histopathology, an appealing direction is to infer spatial spot expression from histology images, enabling the reconstruction of omics profiles from easily accessible pathology data.
As shown in \cref{fig:overview}, each histology image is partitioned into small spots and treated as a unit for spatial gene expression prediction. The goal is to infer the corresponding gene expression of each spot conditioned on its local image content and spatial coordinates.

Several prior works have explored the spatial gene expression prediction problem using different modeling approaches.
Some of them have used pretrained image encoders to directly regress gene expression from local histology patches~\citep{he2020stnet,zeng2022Hist2ST,pang2021histogene,yang2023egn};
some exploited multi-scale feature fusion to capture cross-modal alignment and spatial context~\citep{xie2023bleep, chung2024triplex, chen2025omiclip,wang2024m2ort, chadoutaud2026scellst};
and some went further to leverage generative modeling to better model gene expression distributions and improve prediction~\cite{zhu2025stem, huang2025stflow,fu2025ghist, schroeder2025iscale,liu2025SpatialEx}.

However, these advancements have several inherent limitations that restrict their clinical applicability:
\begin{itemize}[leftmargin=*]
    \item \textbf{Complex pipelines.} These existing approaches rely on complex architectures and multi-stage pipelines, which increase computational cost and introduce additional design and optimization challenges, making them difficult to deploy in resource-constrained clinical environments.
    \item \textbf{Limited gene-panel scalability.} All the previous methods depend on predefined gene panels often limited to a subset of 200 genes, whereas practical clinical applications typically require a coverage of 1000 or more genes; this gap discounts their analysis reliability for downstream tasks.
    \item \textbf{Insufficient interaction modeling.}  Existing methods often fail to capture inter-gene interactions, instead treating genes in a largely independent manner or weakly coupled manner. This lack of structured modeling leads to off-manifold generation, where synthesized gene expression profiles may drift away from the true biological data distribution, especially for near-zero expression values or long-tailed gene distributions.
\end{itemize}
    Therefore, in the sequel, we try to ask and better address the following question:
 \begin{center}
    \emph{Can an efficient generative model support large-panel spatial gene expression prediction solely from histology images?}
 \end{center}

In response, GeneFlow is designed around three key ideas that make flow-based spatial gene expression prediction simple, stable, and scalable.
(1) \textbf{Histology-conditioned gene-expression flow.}
GeneFlow uses pretrained pathology representations to condition a lightweight fusion network that directly predicts target gene expression profiles, bypassing iterative denoising and complex spatial interaction modeling while maintaining efficient histology-to-expression translation~\citep{zhang2025flowpolicy}.
(2) \textbf{ZINB-guided gene token space.}
GeneFlow further maps high-dimensional gene vectors into structured latent gene tokens and initializes the generative process from a ZINB prior, allowing the model to capture cross-gene dependencies under sparsity and overdispersion while remaining scalable to larger gene panels.
Together, these designs enable GeneFlow to model spatial gene expression distributions with fewer architectural assumptions and lower computational cost.
(3) \textbf{Manifold-stable velocity loss.}
To better match the sparse and noisy geometry of ST measurements, GeneFlow predicts the terminal expression endpoint and constructs the velocity from this endpoint, avoiding the unstable normalization that can arise when velocity is estimated from intermediate noisy states~\citep{li2025jit,lu2026pmf}.

\begin{figure}[tb]
	\centering
	\includegraphics[width=.85\linewidth]{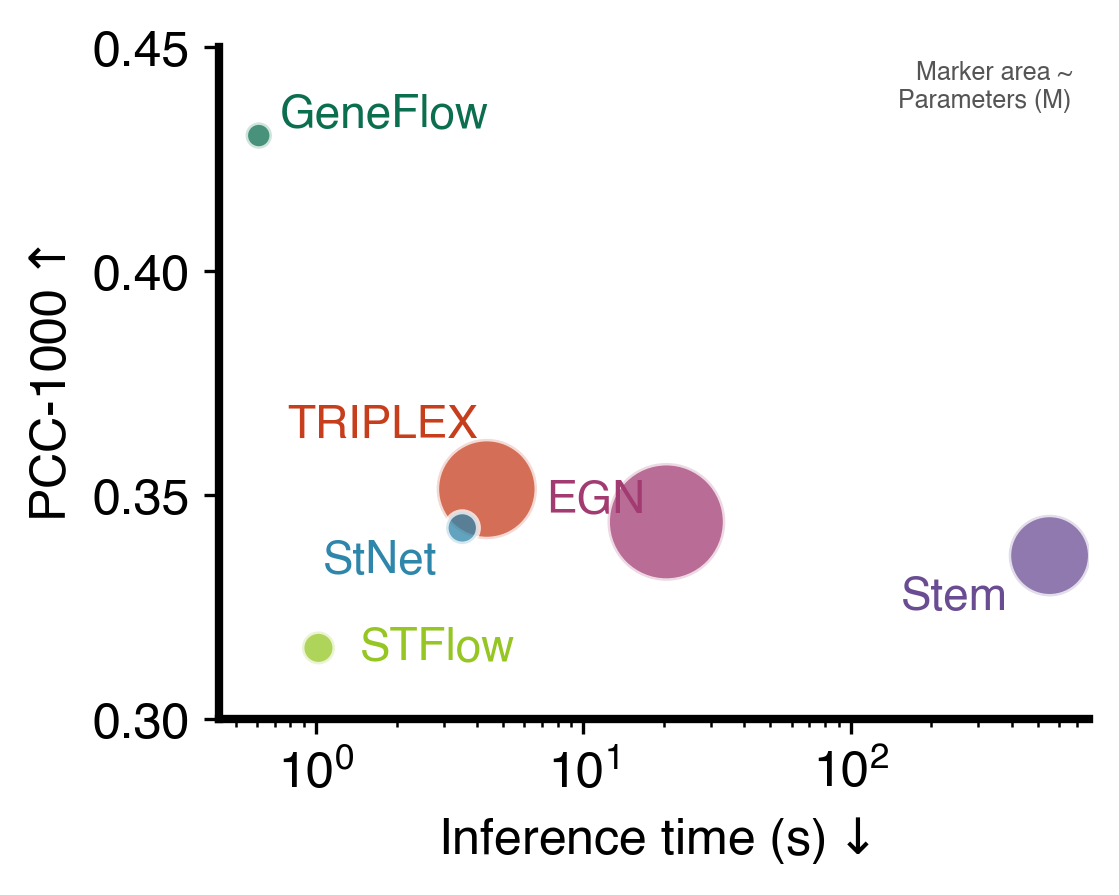}
	\caption{Average PCC-$1000$ vs. inference time across all datasets. GeneFlow delivers high accuracy with low inference time.}
	\label{fig:trade}
    \vspace{-3mm}
\end{figure}

We evaluate GeneFlow on four publicly available datasets spanning different tissue types and sequencing platforms, including kidney, HER2ST, PRAD, and mouse brain.
GeneFlow achieves state-of-the-art performance on gene expression prediction, consistently outperforming existing methods across PCC  metrics, with a 17.1\% improvement on PCC-1000 on the PRAD dataset, highlighting its effectiveness in modeling large-scale gene sets.
GeneFlow also demonstrates improved accuracy in biologically relevant biomarker prediction, suggesting better preservation of meaningful cellular variation.
Moreover,
as shown in \cref{fig:trade},
our GeneFlow also offers orders-of-magnitude reductions in runtime and FLOPs compared to Diffusion-based approaches.
We further analyze the evaluation protocol and show that distance-based metrics such as MSE and MAE are not sufficient to assess prediction quality, as they fail to capture biologically meaningful variation.
We also conduct detailed ablation studies on key design components to validate the robustness and effectiveness of the proposed framework.

To sum up, we have the following summarized contributions:
\begin{itemize}[leftmargin=*]
    \item \textbf{Effective flow-based prediction.}
    We propose GeneFlow, a novel flow-based framework for predicting spatial gene expression from H\&E histology images. It directly models gene expression from image patches without extra multi-stage alignment.
    \item \textbf{Gene-aware multimodal fusion.}
    GeneFlow introduces a lightweight multi-modal fusion architecture to combine image and gene features, enabling efficient prediction over a broader gene set. This design improves flexibility for downstream tasks and supports potential clinical use.
    \item \textbf{Accurate and efficient inference.}
    GeneFlow achieves high-fidelity predictions that are spatially consistent and biologically meaningful over multiple datasets, while also enabling faster inference speed.
\end{itemize}

\section{Related works}
\paragraph{\textbf{Spatial gene expression prediction from histology images.}}
Early deep learning methods for spatial gene expression prediction aim to infer gene expression profiles from histological context~\citep{he2020stnet, pang2021histogene, yang2023egn,zeng2022Hist2ST}. StNet fine-tuned a DenseNet-121 pretrained on ImageNet to directly regress Visium spot gene expression from H\&E image patches~\citep{he2020stnet}.
\citet{xie2023bleep} applies a contrastive learning paradigm: BLEEP maps histology patches and gene-expression profiles into a joint embedding space and enforces biological alignment.
Other approaches emphasize multi-scale and multi-branch architectures to capture tissue context~\citep{schroeder2025iscale, fu2025ghist, liu2025SpatialEx,schmauch2025miso,han2026stamp}.
TRIPLEX uses three parallel encoder branches and fuses features across scales to capture morphology at different resolutions~\citep{chung2024triplex}.
Recent work also explores generative modeling of spatial transcriptomics.
Stem formulates spot-level gene expression prediction as a conditional diffusion process~\citep{zhu2025stem}.
\citet{huang2025stflow} proposes STFlow, a flow-matching generative model that employs a slide-level encoder with local spatial attention to capture cell--cell interactions efficiently.
In contrast, we argue that explicitly modeling spot–spot spatial relations is not always necessary for accurate prediction. Instead, we focus on a simpler and more scalable design that avoids hard-coded priors, enabling efficient extension to larger gene sets and faster inference without sacrificing performance.

\paragraph{\textbf{Flow matching for generative modeling.}}
Flow matching (FM) has recently emerged as an effective paradigm for generative modeling, demonstrating strong performance across diverse domains, including vision, biomolecular modeling, and robotics~\citep{jing2024alphafold,zhang2025flowpolicy,jin2025pyramidal,wang2025vaflow,song2025l1}. Rather than simulating a reverse stochastic process during training, FM learns a time-dependent vector field that transports samples from a simple prior distribution to the target data distribution along a prescribed probability path. This formulation yields a conceptually simple objective and supports efficient, often deterministic, sampling, making it attractive for large-scale generation tasks.
Recent studies have further revisited the parameterization of continuous generative models, suggesting that directly predicting clean signals can simplify the learning objective and improve empirical performance, compared with modeling noise or velocity fields~\citep{li2025jit,lu2026pmf,yu2026scdfm,song2025l1}. Motivated by these advances, we formulate gene expression prediction as a generative process that maps a flexible prior to target expression profiles, which is well suited to large gene panel prediction settings.

\section{Method}
\label{sec:method}

\subsection{Preliminary}
\paragraph{\textbf{Problem formulation.}}
For each spatial spot $s$, we observe an H\&E image patch $I_s \in \mathbb{R}^{\mathcal{H} \times \mathcal{W} \times 3}$ and its coordinates $C_s \in \mathbb{R}^2$, with target gene expression counts $\x_s \in \mathbb{N}_0^{n}$. Given a dataset $\mathcal{D}=\{(I_s, C_s, \x_s)\}_{s=1}^{N}$, the goal is to learn a conditional flow matching model $f_{\theta}$ that predicts $\hat{\x}_s$ from the corresponding $I_s$ under $C_s$:
\begin{equation}
\theta^{\star} = \arg\min_{\theta}\;\frac{1}{N}\sum_{s=1}^{N}\,
\mathcal{L}( \x_s - \hat{\x}_s ),
\end{equation}
where $\mathcal{L}$ is the flow field loss function defined in Sec.\ref{sec:fm}.

\paragraph{\textbf{Flow Matching.}}
Flow Matching (FM) learns a velocity field $\v_\theta$ that transports a prior distribution $p_{\text{prior}}$ to the data distribution $p_{\text{data}}$ by modeling a continuous-time flow. We consider a linear interpolation between a data sample $\x_1 \sim p_{\text{data}}$ and a noise sample $\x_0 \sim p_{\text{prior}}$:
\begin{equation}
\x_t = (1-t)\x_0 + t \x_1, \quad t \in [0,1],
\label{eq:xt}
\end{equation}
which induces the conditional velocity:
\begin{equation}
\v_\theta(\x_t,t) \triangleq \frac{d\x_t}{dt} = \x_1 - \x_0.
\label{eq:v}
\end{equation}

FM parameterizes a neural network $\v_\theta(\x_t,t)$ to approximate this velocity field by minimizing the following objective:
\begin{equation}
\mathcal{L}_{\text{FM}} = \mathbb{E}_{\x_0, \x_1, t} \left\| \v_\theta(\x_t,t) - \v \right\|^2.
\label{eq:fm_loss}
\end{equation}
After training, samples are generated by integrating the learned velocity field from $t=0$ to $t=1$, starting from a noise sample $\x_0 \sim p_{\text{prior}}$:
\begin{equation}
\x_1 = \x_0 + \int_0^1 \v_\theta(\x_t,t)\,dt,
\end{equation}
which is typically approximated using numerical ODE solvers such as Euler or Heun methods.

\subsection{GeneFlow Architecture Fusing Histology and Transcriptomics}
\label{sec:fm}

\paragraph{\textbf{Endpoint-parameterized flow matching.}}
We adopt an endpoint-parameterized variant of flow matching in which
the network directly predicts the terminal sample, while the training
objective remains defined in the velocity space. Given an intermediate
state $\x_t$, the time step $t$, and the histology condition
$\mathbf{c}_I$, the network predicts $\hat{\x}_1=\net_\theta(\x_t,t,\mathbf{c}_I).$
We consider the linear interpolation path $\x_t = (1-t)\x_0 + t\x_1, \qquad t\in[0,1],$
where $\x_0$ is sampled from the prior distribution and $\x_1$ denotes the target gene-expression profile. The corresponding ground-truth conditional velocity is
\begin{equation}
\begin{aligned}
    \v
    =\x_1-\x_0
    &=\frac{(1-t)\x_1-(1-t)\x_0}{1-t}\\
    &=\frac{\x_1-\big((1-t)\x_0+t\x_1\big)}{1-t}\\
    &=\frac{\x_1-\x_t}{1-t}.
\end{aligned}
\end{equation}

Although the network predicts the terminal expression rather than the
velocity directly, its prediction can be converted into an
instantaneous velocity as
\begin{equation}
    \v_\theta(\x_t,t,\mathbf{c}_I)
    =
    \frac{
        \net_\theta(\x_t,t,\mathbf{c}_I)-\x_t
    }{1-t}.
    \label{eq:genetic_v}
\end{equation}

The model is trained using the standard velocity-matching objective:
\begin{equation}
    \mathcal{L}_{\mathrm{FM}}
    = \mathbb{E}_{\x_0,\x_1,t}
    \left[
        \left\|
            \v_\theta(\x_t,t,\mathbf{c}_I)-\v
        \right\|_2^2
    \right]
\end{equation}

This formulation can therefore be interpreted as \emph{x-prediction with a velocity-space loss}: the network output is the clean endpoint $\hat{\x}_1$, whereas supervision is applied after converting the endpoint prediction into the corresponding velocity~\cite{li2025jit}.
For each fixed $(\x_t,t,\mathbf{c}_I)$, this objective and an unweighted endpoint objective share the same optimal endpoint predictor.
However, the velocity-space objective weights endpoint prediction errors by $(1-t)^{-2}$ and therefore places greater emphasis on samples near the terminal time.
Compared with directly predicting the velocity, endpoint parameterization provides the network with an explicit clean-expression prediction target while preserving the standard flow-matching objective and ODE sampling dynamics.

\paragraph{\textbf{Prior distribution.}}
ST data exhibit two key characteristics: high sparsity due to excessive zero counts (i.e., non-activated genes) and overdispersion where the variance exceeds the mean. Standard distributions such as Gaussian or zero fail to capture these properties.
To address this, we adopt a zero-inflated negative binomial (ZINB) distribution  $\mathcal{Z}(\mu, \phi, \pi)$  as the prior $p_{\text{prior}}$:
\begin{equation} \small
p(y \mid \mu, \phi, \pi) =
\begin{cases}
\pi + (1 - \pi)\frac{\Gamma(y+\phi)}{\Gamma(\phi)\, y!}
\left(\frac{\phi}{\phi+\mu}\right)^{\phi}
\left(\frac{\mu}{\phi+\mu}\right)^{y}, & y = 0, \\[8pt]
(1 - \pi)\frac{\Gamma(y+\phi)}{\Gamma(\phi)\, y!}
\left(\frac{\phi}{\phi+\mu}\right)^{\phi}
\left(\frac{\mu}{\phi+\mu}\right)^{y}, & y > 0.
\end{cases}
\end{equation}
Here, $y$ denotes the observed count, $\mu$ is the mean, $\phi$ controls dispersion, and $\pi$ is the zero-inflation probability. The parameter $\phi$ allows the negative binomial component to capture overdispersion beyond the Poisson assumption. This provides a more accurate statistical description of gene expression patterns and improves the modeling of realistic count data during training.

\paragraph{\textbf{Sampling.}}
We start from an initial expression sample $\x_0$ drawn from the ZINB distribution $\mathcal{Z}(\mu, \phi, \pi)$ and iteratively refine it through the learned flow model. At each step, given the intermediate state $\x_t$, the network $\net_\theta(\x_t, t)$ predicts the terminal expression, which is then used to update the trajectory according to the flow formulation. This process is repeated across time steps and produces the final gene expression at $t=1$.

\paragraph{\textbf{Histological tokenizer.}}
Instead of training a pathology encoder from scratch, we directly leverage pretrained foundation models to extract histological representations.
These models are trained on large-scale pathology datasets and provide general-purpose features with strong transferability, which is particularly beneficial in ST settings where labeled data are limited and distribution shifts across slides are common. Such representations also help mitigate batch effects by capturing more stable morphological patterns.
Formally, we denote $f_{\text{H}}(\cdot)$ as a pathology foundation model that maps an input H\&E image patch $I_i$ to a pathology embedding $E_i$, i.e.,
$$\{E_0, \cdots, E_N\} = f_{\text{H}}(\{I_0, \cdots, I_N\}).$$
In practice, we follow Stem~\citep{zhu2025stem} setting to obtain patch-level embeddings produced by UNI~\cite{chen2024towards} and CONCH~\cite{huang2023visual}, and project them to the generative model hidden space for subsequent processing.
To further improve robustness and reduce overfitting, we also adopt lightweight data augmentation strategies, including rotations and flipping, without altering the underlying tissue structure. The resulting histological embeddings are then used as conditional inputs for downstream gene expression modeling.

\paragraph{\textbf{Gene tokenizer.}}
We assume that local morphological context in ST is implicitly provided by the pretrained histological representations for each spatial spot; therefore, to maintain a simple and efficient design, we do not explicitly encode the spatial coordinates or inter-spot relations.
Given a gene expression vector $x^{(k)} \in \mathbb{R}^{G}$ for each spot $k$, we first project it into a latent token space using the gene tokenizer. Formally, the tokenization process can be written as
$$ Z^{(k)} = \phi\!\left(x^{(k)} W + b\right) + P, $$
where $W \in \mathbb{R}^{G \times (h d)}$ and $b \in \mathbb{R}^{h d}$ denote the learnable projection parameters, $\phi(\cdot)$ is a non-linear activation function, and $P \in \mathbb{R}^{h \times d}$ is a learnable gene token-type embedding. The output is then reshaped into $Z^{(k)} \in \mathbb{R}^{h \times d}$ for each spot, where $h$ denotes the number of gene module tokens and $d$ is the token dimension.
By compressing high-dimensional gene expression into a fixed number of gene module tokens, the model captures shared structure among genes while maintaining a compact and computationally efficient representation.
In addition, the modular tokenization provides a more interpretable latent space, as each token can capture a relatively coherent functional gene program rather than an entangled mixture of unrelated signals.
As we show in \cref{fig:enrichment}, these learned tokens tend to align with biologically meaningful and functionally distinct pathways.

\begin{figure}[tb]
	\centering
    \includegraphics[width=\linewidth]{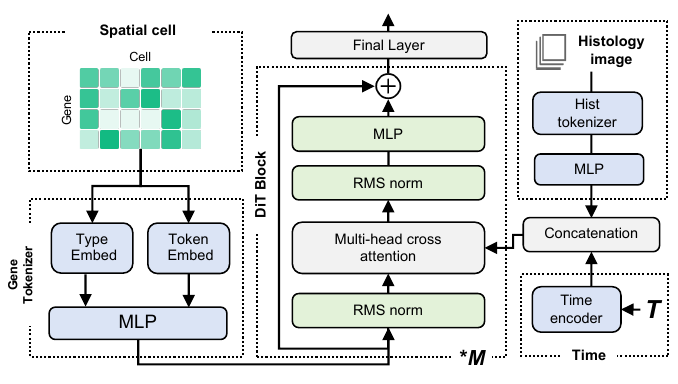}
	\caption{Visualization of GeneFlow architecture.    }
	\label{fig:architecture}
\end{figure}

\paragraph{\textbf{Overall fusion architecture}}
Our GeneFlow network is built on a DiT-based transformer~\cite{dit} that parameterizes a flow matching model.
Gene expression inputs are encoded using a gene tokenizer, resulting in a sequence of hidden vectors with $D$ dimensions. Time information is encoded using a standard 256-dimensional sinusoidal embedding followed by a 2-layer MLP. For histology, each patch is represented by a single embedding extracted from pretrained pathology foundation models, either via special tokens or pooling; when multiple models are used, their embeddings are concatenated and projected through a 2-layer MLP to obtain a unified conditioning vector. This conditioning vector, together with the time embedding, is injected into the transformer blocks via adaLN-Zero modulation. The model consists of 12 DiT blocks with 6-head attention, followed by a final adaLN and linear layer to predict the flow field. An overview of the architecture is presented in \cref{fig:architecture}.

\paragraph{\textbf{Some useful tricks.}}
1. Parameter-free RMSNorm. We apply RMS normalization~\citep{zhang2019rmsnorm} without learnable parameters throughout the DiT network.
Given the high sparsity and large dynamic range of gene expression data, this design provides stable feature scaling across spots and genes without introducing additional parameters. It is inserted after the input embedding, before each attention and MLP block, and before the final prediction head, ensuring stable feature scaling with zero parameter overhead.
2. ReLU$^2$ activation. We replace GELU with a squared ReLU activation~\citep{zhang2024relu2} in MLP blocks, i.e., \texttt{F.relu(x).square()}. This activation better matches the non-negative and sparse nature of gene expression signals, promoting sparse responses without sacrificing representational capacity.
3. QK normalization. Although RoPE does not provide noticeable gains in our setting, we retain query--key normalization and apply it directly as $q, k = \mathrm{norm}(q), \mathrm{norm}(k)$. This stabilizes attention computation and removes the need for additional techniques.

\subsection{The generalized manifold perspective of our formulation}
\label{sec:manifold}

\paragraph{\textbf{Boundary cases.}}
When $t\to0$, the interpolated state approaches the prior sample, i.e., $\x_t\to\x_0$, and the predicted endpoint $\hat{\x}_1=\net_\theta(\x_t,t,\mathbf{c}_I)$ provides an explicit target on the gene-expression data manifold.
The resulting velocity $\v_\theta=(\hat{\x}_1-\x_t)/(1-t)$ therefore points from the prior toward a predicted clean expression profile. When $t\to1$, both $\x_t$ and the ideally predicted $\hat{\x}_1$ approach the clean data $\x_1$, so the displacement $\hat{\x}_1-\x_t$ connects two states close to the data manifold.
Meanwhile, the direction of the predicted velocity is still anchored by an explicit terminal expression estimate rather than an unconstrained velocity output.

\paragraph{\textbf{General case.}}
For $t\in(0,1)$, the interpolated state $\x_t$ may lie outside the low-dimensional and highly non-uniform manifold of observed gene-expression profiles. GeneFlow first maps this intermediate state and its histology condition to a clean endpoint estimate $\hat{\x}_1$, which is directly supervised by real expression profiles, and then constructs the velocity from the displacement between $\x_t$ and $\hat{\x}_1$. This endpoint-oriented parameterization encourages the predicted flow to point toward biologically plausible regions of the expression space, whereas direct velocity prediction does not explicitly associate its output with a point on the data manifold. Our empirical results suggest that it provides a useful manifold-anchored inductive bias for sparse, heavy-tailed, and technically noisy ST data.

\section{Experiment}
\label{sec:experiment}
In this section, we evaluate GeneFlow on spatial gene expression prediction and conduct ablation studies across four datasets, comparing it against six representative baselines.
To make the analysis explicit, we gradually answer the following research questions:
\begin{itemize}[leftmargin=*]
    \item \textbf{RQ1:} Is GeneFlow truly effective in ST benchmarks?
    \item \textbf{RQ2:} Are GeneFlow predictions biologically meaningful?
    \item \textbf{RQ3:} Is GeneFlow sufficiently efficient and scalable?
    \item \textbf{RQ4:} What drives the performance of GeneFlow?
\end{itemize}

\subsection{Training Details}
\label{sec:training_details}

\paragraph{\textbf{Datasets}}
We evaluated our method on four ST datasets from the HEST-1k database~\citep{jaume2024hest}, covering multiple organs, species, and disease conditions~\citep{andersson2021spatial, lake2023atlas,vicari2024spatial,erickson2022spatially}.
All datasets were generated using spot-based ST platforms and provide paired histology images and spatially resolved gene expression profiles, as summarized in \cref{tab:dataset}.
Following the slide-level holdout protocol~\cite{zhu2025stem}, we reserve one predefined slide from each dataset for evaluation and use all remaining slides for training.

\begin{table*}[tb]
  \centering
  \setlength{\tabcolsep}{6.5pt}
  \caption{The results of gene expression prediction across spatial omics datasets. \textbf{Bold} indicates the best performance, \underline{underlined} indicates the second-best performance.}
    \begin{tabular}{llcccccccc}
      \toprule[1pt]
      \multirow{2}{*}{Dataset}
      & \multirow{2}{*}{Metric}
      & StNet & EGN & BLEEP & TRIPLEX & Stem & STFlow
      & \multirow{2}{*}{GeneFlow}
      & \multirow{2}{*}{{\small\textcolor{gray}{$\Delta(\uparrow)$}}} \\
      & &
      {\scriptsize\textcolor{gray}{(Nat. BE, 2020)}} &
      {\scriptsize\textcolor{gray}{(CVPR, 2023)}} &
      {\scriptsize\textcolor{gray}{(NeurIPS, 2023)}} &
      {\scriptsize\textcolor{gray}{(CVPR, 2024)}} &
      {\scriptsize\textcolor{gray}{(ICLR, 2025)}} &
      {\scriptsize\textcolor{gray}{(ICML, 2025)}} &
      & \\
      \midrule

    \multirow{4}{*}{\textit{Kidney}}
    & PCC-10   & 0.4337 & 0.3039 & 0.2779 & 0.4641 & \underline{0.4662} & 0.3301 & \textbf{0.5837}{\footnotesize\textcolor{gray}{$\pm$0.0012}} & {\small\textcolor{gray}{+25.21\%}} \\
    & PCC-50   & 0.3680 & 0.1807 & 0.2165 & 0.3823 & \underline{0.4121} & 0.2544 & \textbf{0.5345}{\footnotesize\textcolor{gray}{$\pm$0.0023}} & {\small\textcolor{gray}{+29.71\%}} \\
    & PCC-200  & 0.2936 & 0.1562 & 0.1632 & 0.2769 & \underline{0.3200} & 0.1824 & \textbf{0.4545}{\footnotesize\textcolor{gray}{$\pm$0.0013}} & {\small\textcolor{gray}{+42.03\%}} \\
    & PCC-1000 & 0.1604 & 0.1473 & 0.0532 & \underline{0.1923} & 0.1752 & 0.0800 & \textbf{0.3159}{\footnotesize\textcolor{gray}{$\pm$0.0013}} & {\small\textcolor{gray}{+64.27\%}} \\

    \midrule

    \multirow{4}{*}{\textit{Mouse Brain}}
    & PCC-10   & \underline{0.5412} & 0.4687 & 0.3800 & 0.5373 & 0.5393 & 0.4757 & \textbf{0.5886}{\footnotesize\textcolor{gray}{$\pm$0.0005}} & {\small\textcolor{gray}{+8.76\%}} \\
    & PCC-50   & \underline{0.4850} & 0.2400 & 0.3010 & 0.4708 & 0.4644 & 0.4187 & \textbf{0.5281}{\footnotesize\textcolor{gray}{$\pm$0.0001}} & {\small\textcolor{gray}{+8.88\%}} \\
    & PCC-200  & \underline{0.3976} & 0.2413 & 0.2091 & 0.3778 & 0.3693 & 0.3355 & \textbf{0.4426}{\footnotesize\textcolor{gray}{$\pm$0.0003}} & {\small\textcolor{gray}{+11.33\%}} \\
    & PCC-1000 & 0.2369 & 0.2442 & 0.0798 & \underline{0.2827} & 0.2147 & 0.1935 & \textbf{0.2882}{\footnotesize\textcolor{gray}{$\pm$0.0005}} & {\small\textcolor{gray}{+1.95\%}} \\

    \midrule

    \multirow{4}{*}{\textit{HER2ST}}
    & PCC-10   & 0.8136 & 0.8203 & 0.4792 & 0.7722 & 0.8418 & \underline{0.8591} & \textbf{0.8664}{\footnotesize\textcolor{gray}{$\pm$0.0012}} & {\small\textcolor{gray}{+0.84\%}} \\
    & PCC-50   & 0.7792 & 0.7089 & 0.4056 & 0.7208 & 0.7934 & \underline{0.8210} & \textbf{0.8323}{\footnotesize\textcolor{gray}{$\pm$0.0012}} & {\small\textcolor{gray}{+1.37\%}} \\
    & PCC-200  & 0.7407 & 0.6652 & 0.2875 & 0.6346 & 0.7388 & \underline{0.7850} & \textbf{0.7957}{\footnotesize\textcolor{gray}{$\pm$0.0013}} & {\small\textcolor{gray}{+1.36\%}} \\
    & PCC-1000 & 0.5842 & 0.6419 & 0.0347 & 0.5520 & 0.5654 & \underline{0.6564} & \textbf{0.6738}{\footnotesize\textcolor{gray}{$\pm$0.0010}} & {\small\textcolor{gray}{+2.66\%}} \\

    \midrule

    \multirow{4}{*}{\textit{PRAD Visium}}
    & PCC-10   & \underline{0.6943} & 0.5357 & 0.4603 & 0.6703 & 0.6780 & 0.5844 & \textbf{0.7269}{\footnotesize\textcolor{gray}{$\pm$0.0047}} & {\small\textcolor{gray}{+4.69\%}} \\
    & PCC-50   & \underline{0.6472} & 0.3538 & 0.4000 & 0.6113 & 0.6342 & 0.5259 & \textbf{0.6756}{\footnotesize\textcolor{gray}{$\pm$0.0125}} & {\small\textcolor{gray}{+4.39\%}} \\
    & PCC-200  & \underline{0.5517} & 0.3762 & 0.3341 & 0.5049 & 0.5477 & 0.4522 & \textbf{0.5895}{\footnotesize\textcolor{gray}{$\pm$0.0145}} & {\small\textcolor{gray}{+6.86\%}} \\
    & PCC-1000 & 0.3895 & 0.3430 & 0.2199 & 0.3787 & \underline{0.3910} & 0.3340 & \textbf{0.4307}{\footnotesize\textcolor{gray}{$\pm$0.0009}} & {\small\textcolor{gray}{+10.16\%}} \\

    \bottomrule[1pt]
  \end{tabular}
  \label{tab:spatial_gene_pred}
\end{table*}

\paragraph{\textbf{Experiment setup.}}
We compare our method with several state-of-the-art approaches for spatial gene expression prediction, including StNet~\citep{he2020stnet}, EGN~\citep{yang2023egn}, BLEEP~\citep{xie2023bleep}, TRIPLEX~\citep{chung2024triplex}, Stem~\citep{zhu2025stem}, and STFlow~\citep{huang2025stflow}.
All baselines are implemented using their official codebases and original training configurations.
To maintain training stability, we clip the gradient norm to 1.0 at each iteration. The learning rate is selected from the range $\{1\mathrm{e}$-$3, 5\mathrm{e}$-$4, 1\mathrm{e}$-$4\}$, and we set the default value to $1\mathrm{e}$-$4$.
All training and inference are performed in the original gene-count space, ensuring that the ground-truth expression and the ZINB prior lie in the same numerical space.
In addition, all parameters of the pre-trained pathology foundation models are frozen during training.
A 224×224 image patch centered at each spatial spot is cropped as input.
Following the common gene selection protocol~\citep{he2020stnet}, we select the top 1000 genes on all training slides from the intersection of highly expressed and highly variant genes.
We employ the AdamW optimizer in our experiments for GeneFlow~\citep{loshchilov2018decoupled}.
All experiments in this work are conducted on a Linux server equipped with an AMD EPYC 7H12 64-Core Processor. The server is configured with NVIDIA RTX A6000 GPUs for model training and inference.

\paragraph{\textbf{Evaluation metrics.}}
We evaluate prediction performance using both correlation-based and error-based metrics.
Specifically, PCC-$k$ denotes the average Pearson correlation coefficient over the top-$k$ best-predicted genes, reflecting gene-wise prediction quality ($k = 10, 50, 200, 1000$).
It is computed as
\begin{equation}
    \operatorname{PCC}\text{-}k = \frac{1}{k} \sum_{g \in \text{Top-k}}
    \frac{\operatorname{Cov}(\x_g, \hat{\x}_g)}{\sqrt{\operatorname{Var}(\x_g)\cdot \operatorname{Var}(\hat{\x}_g)}},
\end{equation}
where $\x_g$ and $\hat{\x}_g$ are the true and predicted expression values of gene $g$, and $\operatorname{Cov}(\cdot)$ and $\operatorname{Var}(\cdot)$ denote covariance and variance.
Additionally, MSE and MAE measure global prediction error across all genes and spatial spots.

\begin{figure*}[tb]
    \centering
    \includegraphics[width=.95\linewidth]{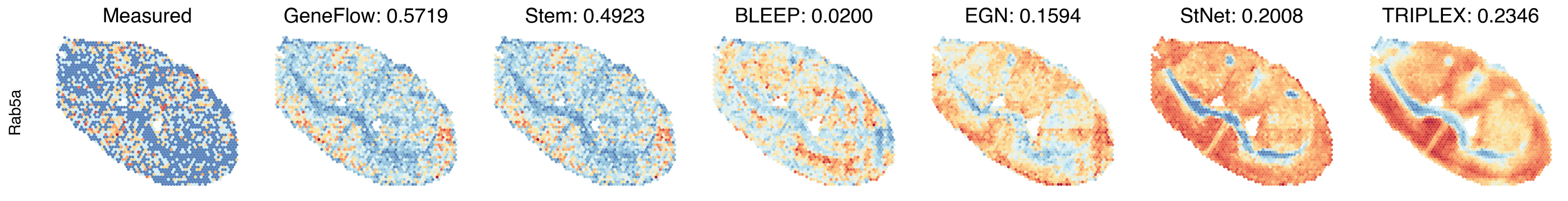}
    \includegraphics[width=.95\linewidth]{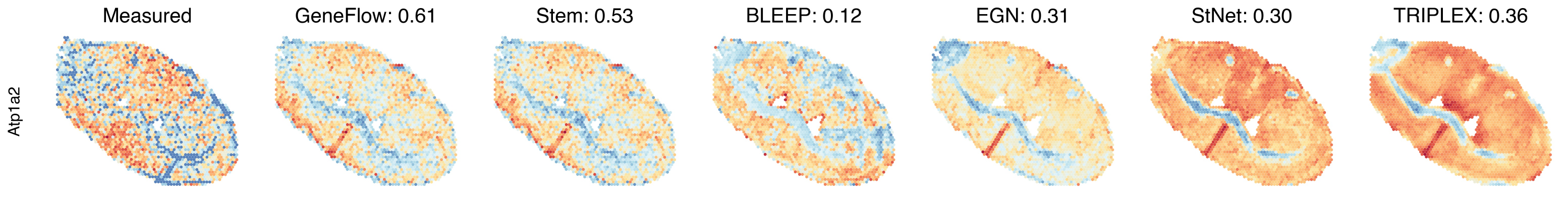}
    \caption{Visualization of biomarker gene expression with the corresponding PCC values. The color scale ranges from blue (low expression) to red (high expression).
    Predicted expression of \textit{Rab5a} and \textit{Atp1a2} on the Mouse Brain tissue slide NCBI667.
    }
    \label{fig:biomarker}
\end{figure*}

\subsection{RQ1: Robust prediction across diverse ST datasets}

We first examine whether GeneFlow can accurately predict spatial gene expression across diverse tissues and disease contexts.
The results in \cref{tab:spatial_gene_pred} show that GeneFlow consistently achieves the highest PCC scores across all datasets, indicating more accurate modeling of spatial gene expression patterns.
On both cancer datasets (HER2ST and PRAD), GeneFlow improves PCC-200 by 1.4\% and 9.8\% over the best baselines, reflecting its ability to capture cell state variations across spatial contexts.
Notably, the improvements on PCC-1000 are even more pronounced, reaching up to 2.7\% on HER2ST and 17.1\% on PRAD.
This result demonstrates that the advantage of GeneFlow is maintained as the gene panel grows, supporting the effectiveness of the gene tokenizer in modeling cross-gene dependencies within large expression panels.
GeneFlow also surpasses all competing methods in terms of PCC on the Kidney and Mouse Brain datasets, demonstrating that its predictive performance generalizes beyond cancer tissues to both pathological and normal biological settings.
We provide a more detailed discussion in Sec. 4.5 of the relationship between PCC and absolute-error metrics (\textit{e.g.}, MAE and MSE), including their potentially differing performance trends~\citep{arcinstitute2025vcc}.
More detailed ablation studies are shown in \cref{tab:component_ablation}.
These improvements are primarily attributed to the genetic flow matching mechanism of GeneFlow, which explicitly aligns predicted and observed gene expression distributions in a spatially coherent manner.

\subsection{RQ2: Preservation of biologically meaningful spatial patterns}

One of the important applications of spatial gene expression prediction is to understand tissue-specific biological patterns in relation to morphology.
To assess these properties, we visualize the slide-level spatial expression maps predicted by GeneFlow and the competing methods, using the corresponding measured expression maps as ground truth. Specifically, we select two representative biomarker genes, \textit{Rab5a} and \textit{Atp1a2}, from the Mouse Brain dataset.

As shown in \cref{fig:biomarker,fig:biomarker_2},  GeneFlow produces predictions that closely align with the observed spatial distribution of these genes, preserving both high-expression regions and spatial boundaries. In contrast, competing methods exhibit clear limitations: BLEEP and EGN generate nearly uniform maps with negligible dynamic range, failing to capture spatial heterogeneity; and Stem still suffers from mild oversmoothing, reducing contrast between expression domains.
These qualitative observations are consistent with the quantitative results and further demonstrate that GeneFlow does not merely improve numerical prediction accuracy: it also preserves biologically meaningful spatial expression patterns that reflect the underlying tissue organization.

\begin{table}[tb]
\centering
 \setlength{\tabcolsep}{9pt}
\caption{Computational efficiency comparison using \texttt{torch.profiler}. }
\label{tab:param}
\begin{tabular}{lccc}
\toprule
Model   & Param (M) & GFLOPs & Time (s) \\
\midrule
StNet   & 8.0         & 5.57 & 3.52 \\
EGN     & 231       & 17.66 & 20.42 \\
BLEEP   & 12.5      & 3.63 & 48.3     \\
TRIPLEX & 85.1      & 33.56 & 4.35 \\
STFlow  & 8.1       & 53.35 & 1.02 \\
Stem    & 35.0       & 42.79 & 554 \\
\rowcolor[HTML]{EFEFEF}
GeneFlow   & 7.3 & 0.31 & 0.61 \\
\bottomrule
\end{tabular}
\vspace{-3mm}
\end{table}

\begin{figure}[tb]
    \centering
    \includegraphics[width=.95\linewidth]{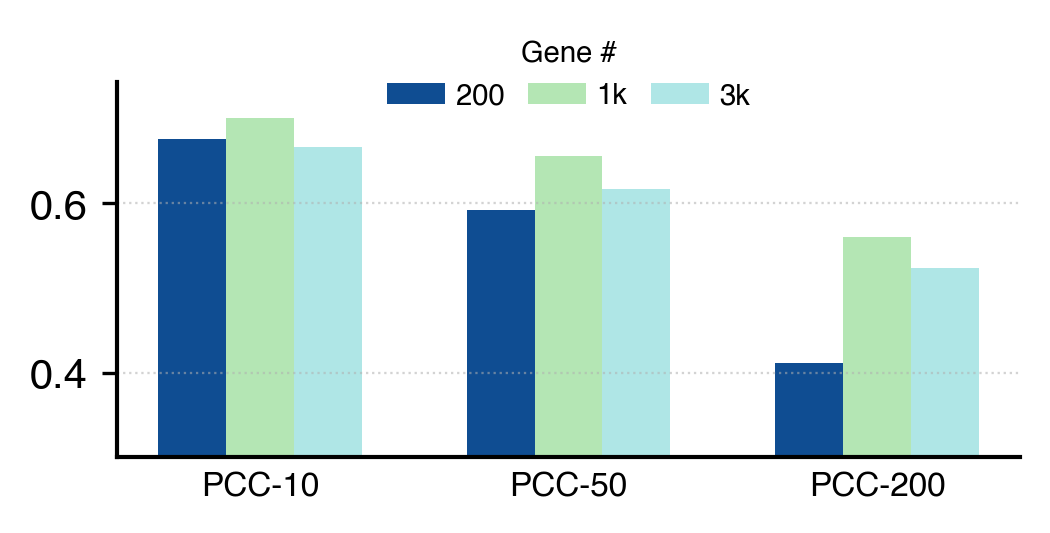}
    \caption{Results of the ablation study on the scalability of GeneFlow to large gene sets. Here, \emph{gene num} represents the number of predicted genes provided to the model.}
    \label{fig:ablation_gene_num}
\end{figure}

\subsection{RQ3: Efficiency and scalability}

\paragraph{\textbf{Computational efficiency of GeneFlow}}
We assess the computational efficiency of GeneFlow relative to other baselines by measuring Floating Point Operations (GFLOPs per sample) and sampling time using \texttt{torch.profiler}. As shown in \cref{tab:param}, GeneFlow achieves the lowest computational cost (0.31 GFLOPs) for the trainable prediction module among all methods, and also the shortest sampling time (0.61 s per sample). Compared with STFlow (53.35 GFLOPs) and Stem (42.79 GFLOPs), GeneFlow reduces computational cost by more than one order of magnitude while also being significantly faster in inference.
GeneFlow offers a favorable balance between prediction quality, inference speed, and computational cost, making it suitable for time-sensitive spatial omics analysis and downstream clinical decision support.

\paragraph{\textbf{Scaling to larger gene sets}}
In this ablation experiment, we evaluate the scalability of GeneFlow by testing on a large gene set. We select 3000 Differentially Expressed Genes (DEGs) from the PRAD dataset.
The evaluation result is shown in \cref{fig:ablation_gene_num}. Despite scaling to 3000 gene-level targets, GeneFlow maintains relatively strong prediction quality across correlation-based metrics, indicating that performance does not sharply degrade as the predicted gene panel grows. Notably, GeneFlow still maintains competitive PCC-$k$ scores in this setting, suggesting that it continues to capture complex spatial heterogeneity when jointly predicting many genes.

\begin{figure}[tb]
    \centering
    \includegraphics[width=\linewidth]{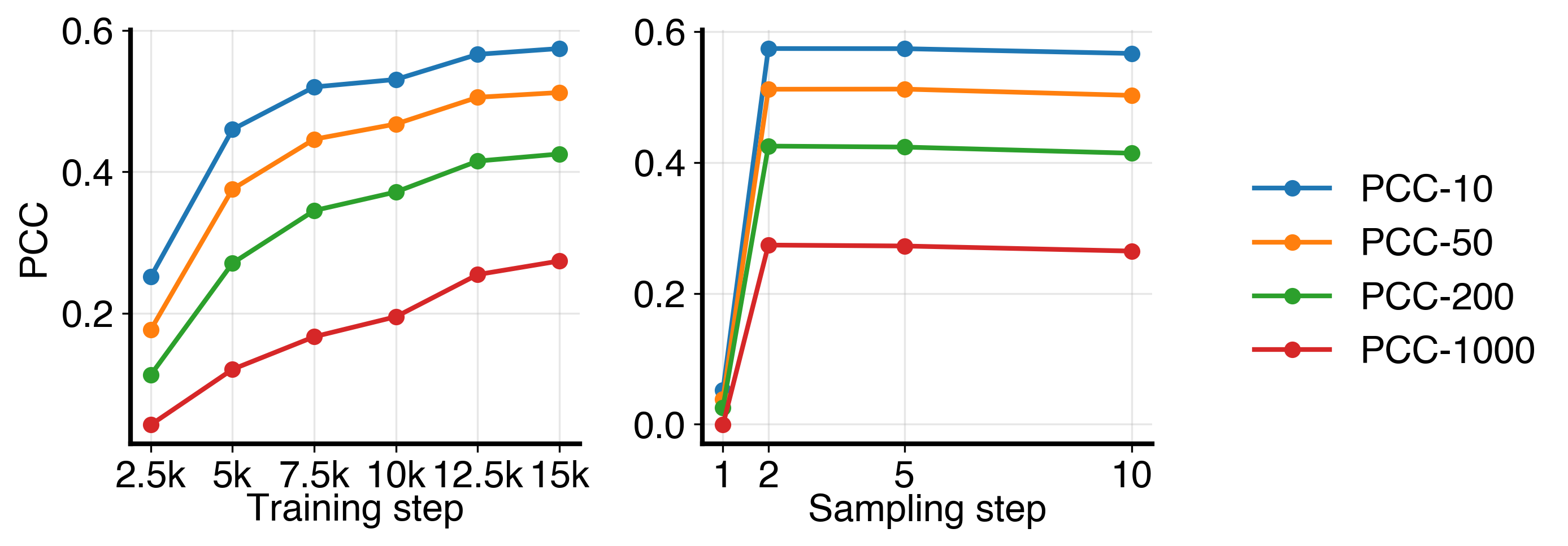}
    \caption{Effect of Training and Sampling Steps.}
    \label{fig:pcc_over_steps}
    \vspace{-3mm}
\end{figure}

\paragraph{\textbf{Effect of training and sampling steps.}}
As shown in \cref{fig:pcc_over_steps}, training steps are positively correlated with PCC across all scales. The most significant PCC gains occur from 2.5k to 7.5k steps, after which improvement saturates as the model converges. The best performance is achieved at 15k steps.
Unlike training steps, sampling steps show a distinct pattern: PCC jumps sharply from 1 to 2 steps. As 1 step fails to capture spot expression, while 2 steps reduce sampling randomness. Performance stabilizes beyond 2 steps, indicating 2–5 steps are sufficient to balance prediction accuracy and computational efficiency.

\subsection{RQ4: Further Analysis}
\label{sec:further_analysis}

\begin{table}[tb]
    \centering
    \caption{Ablation study of GeneFlow.}
    \label{tab:component_ablation}
    \resizebox{\linewidth}{!}{
    \begin{tabular}{lcccc}
        \toprule
        & PCC-10 & PCC-50 & PCC-200 & PCC-1000 \\ \midrule
        GeneFlow & \textbf{0.5888} & \textbf{0.5277} & \textbf{0.4424} & \textbf{0.2874} \\
        ~~w/o gene tokenizer        & 0.5070 & 0.4521 & 0.3880 & 0.1997 \\
        ~~w/o~ CONCH & 0.3024 & 0.2379 & 0.1686 & 0.0645 \\
        ~~w/o~ UNI & 0.2478 & 0.1865 & 0.1312 & 0.0374 \\
        ~~w/~  $\v$-pred / $\x$-loss & 0.5342 & 0.4731 & 0.3848 & 0.2642\\
        ~~w/~ $\v$-pred / $\v$-loss & 0.5214 & 0.4608 & 0.3824 & 0.2498 \\
        ~~w/~ $\x$-pred / $\x$-loss & 0.5707 & 0.5069 & 0.4096 & 0.2613 \\
        \bottomrule
    \end{tabular}
    }
    \vspace{-3mm}
\end{table}

\begin{figure}[tb]
    \centering
    \includegraphics[width=\linewidth]{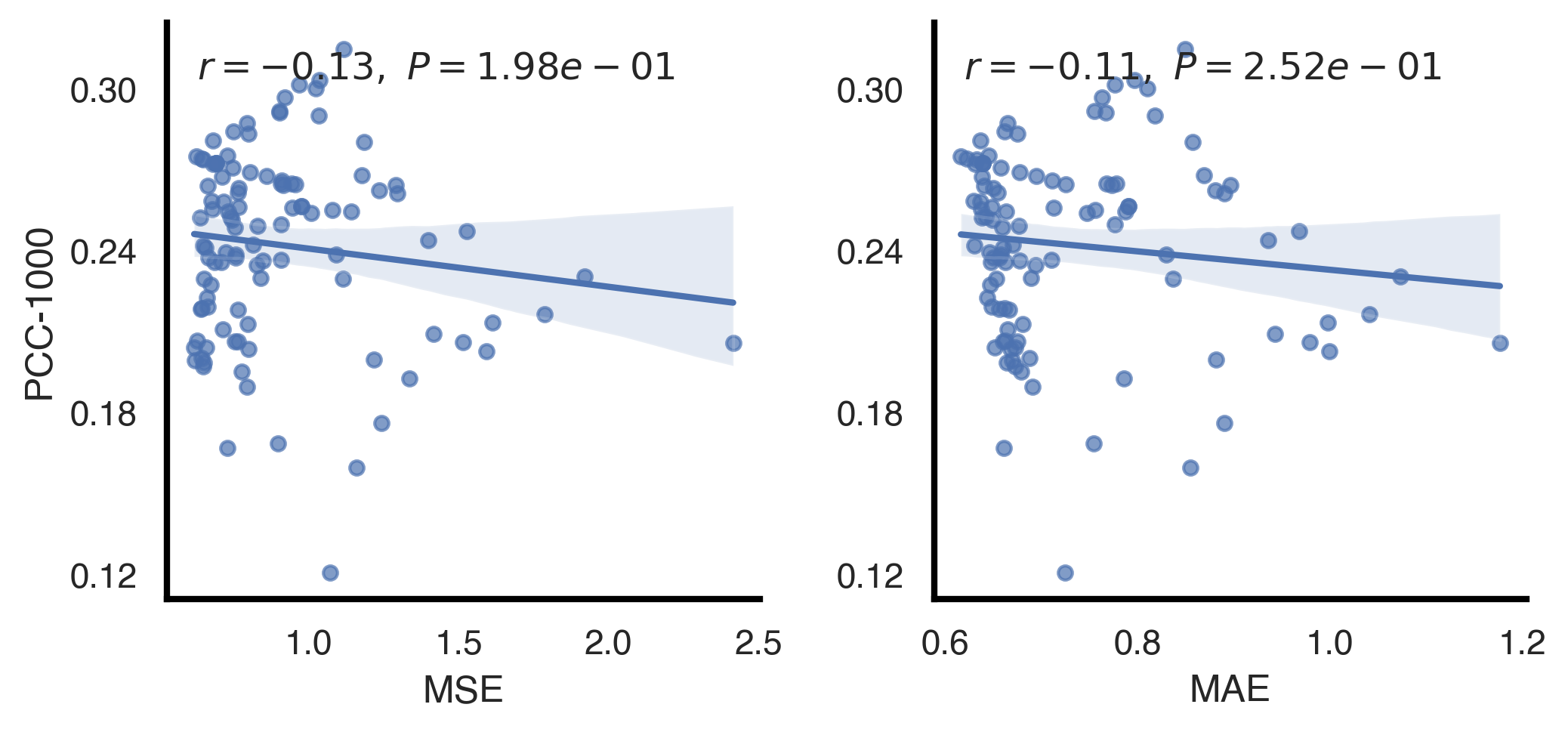}
    \caption{Correlation between PCC and MSE/MAE}
    \label{fig:pcc_vs_error}
    \vspace{-3mm}
\end{figure}

\paragraph{\textbf{Correlation vs. error metrics analysis.}}
We further analyze the relationship between PCC and error-based metrics (MSE, MAE) using the scatter plots presented in \cref{fig:pcc_vs_error}. Our analysis reveals that PCC exhibits only a weak negative correlation with both MSE and MAE, with correlation coefficients of $r=-0.13$ and $r=-0.11$, respectively. This weak linear relationship suggests that minimizing reconstruction error does not guarantee maximized correlation, highlighting the distinct nature of these metrics.
Furthermore, we observe substantial vertical dispersion in \cref{fig:pcc_vs_error}, particularly within the low-error regime. Specifically, predictions achieving comparable error values often display significantly divergent PCC scores. This pronounced variability indicates that model performance in terms of reconstruction error does not linearly translate to predictive correlation. Consequently, optimizing solely for lower MSE or MAE may not reliably enhance the biological relevance.

\begin{table}[tb]
\centering
 \setlength{\tabcolsep}{7pt}
\caption{Performance under different token dim and gene num ratios $D/G$}.
\label{tab:token_gene_ratio}
\begin{tabular}{crcccc}
\toprule
Ratio $D/G$   & PCC-10  & PCC-50  & PCC-200 & PCC-1000   \\
\midrule
1:5                 & 0.5789  & 0.5128  & 0.4217  & 0.2604   \\
1:10                & 0.5888  & 0.5277  & 0.4424  & 0.2874   \\
1:20               & 0.5563  & 0.4897  & 0.3899  & 0.2105  \\
1:50                & 0.5755  & 0.5182  & 0.4306  & 0.2708  \\
1:100              & 0.5746  & 0.5169  & 0.4233  & 0.2678  \\
\bottomrule
\end{tabular}
\vspace{-3mm}
\end{table}

\paragraph{\textbf{Effect of Gene Tokenizer Dimensions.}}\
\cref{tab:hidden_dim,tab:token_gene_ratio} examine how hidden dimension and gene-token compression ratio affect performance and model complexity. Increasing the hidden dimension from 12 to 24 consistently improves PCC, with the best results at 24. Further increases add parameters but bring no stable gains and may slightly hurt performance. The gene-token dimension shows a trade-off between accuracy and efficiency: lower compression preserves more information, while higher compression reduces parameters and computation. However, excessive compression (e.g., $D/G=1{:}20$) creates an information bottleneck and degrades performance. Thus, the compression ratio can be chosen based on application needs.

\paragraph{\textbf{Effect of prior distribution.}}
\cref{tab:prior_zinb} evaluates different prior distributions. ZINB consistently outperforms the Zero prior, especially on larger gene sets (e.g., PCC-1000: 0.2739 vs. 0.1763), by better capturing sparse gene expression patterns. Compared with the Gaussian prior, ZINB achieves comparable performance on highly correlated genes and stronger performance on broader gene sets, suggesting that modeling zero inflation improves the recovery of weak gene signals. These results indicate that ZINB provides a more suitable prior for ST data by preserving sparse expression characteristics and gene-level variation.

\begin{table}[tb]
    \centering
    \caption{Performance under different prior types and ZINB parameter settings}
    \label{tab:prior_zinb}
    \resizebox{.95\linewidth}{!}{
    \begin{tabular}{llcccc}
    \toprule
    \multicolumn{2}{c}{Prior Type} & PCC-10 & PCC-50 & PCC-200 & PCC-1000 \\ \midrule
    \multicolumn{2}{c}{Zero}       & 0.5108 & 0.4417 & 0.3411 & 0.1763 \\
    \multicolumn{2}{c}{Gaussian}   & 0.5641 & 0.5023 & 0.4137 & 0.2625 \\ \midrule
    \multirow{9}{*}{ZINB} & (0.1, 1) & 0.5743 & 0.5121 & 0.4252 & 0.2739 \\
                          & (0.1, 2) & 0.5666 & 0.5043 & 0.4144 & 0.2674 \\
                          & (0.1, 4) & 0.5432 & 0.4848 & 0.3936 & 0.2374 \\
                          & (0.2, 1) & 0.5535 & 0.4938 & 0.4056 & 0.2582 \\
                          & (0.2, 2) & 0.5551 & 0.4976 & 0.4110 & 0.2552 \\
                          & (0.2, 4) & 0.5512 & 0.4925 & 0.4008 & 0.2491 \\
                          & (0.4, 1) & 0.5632 & 0.4966 & 0.4061 & 0.2563 \\
                          & (0.4, 2) & 0.5461 & 0.4845 & 0.4006 & 0.2525 \\
                          & (0.4, 4) & 0.5397 & 0.4813 & 0.3921 & 0.2385 \\
    \bottomrule
    \end{tabular}
    }
    \vspace{-5mm}
\end{table}

\section{Conclusion}
In this work, we propose GeneFlow, a novel framework for spatially resolved gene expression inference from H\&E-stained histology images. By combining flow-based generative modeling with structured multi-modal fusion, GeneFlow provides a simple yet effective way to capture both morphological cues and cross-gene dependencies in a unified representation space. As a result, GeneFlow generates highly accurate and biologically meaningful predictions for unseen histology images at test time and achieves state-of-the-art performance across multiple datasets and evaluation metrics. Its strong accuracy, efficiency, and scalability suggest that simple generative designs can serve as practical and competitive solutions for large-scale spatial gene expression prediction.

\section{Limitations and Ethical Considerations }
Our results are bounded by several limitations.
First, its prediction accuracy is constrained by the quality and representation capacity of the pathology foundation model.
Second, GeneFlow relies primarily on paired H\&E and ST data for supervision and does not explicitly incorporate external biological knowledge, which may limit its generalizability in settings where paired training data are scarce.
For future work, it would be interesting to extend GeneFlow by incorporating additional modalities, such as language information.
We conduct secondary analysis of publicly available, de-identified human and animal datasets without collecting new specimens or accessing personally identifiable information. We position this work for research use and recommend that any deployment follow institutional biosafety review and applicable regulations.

\section{Generative AI Usage}
We used generative AI tools to assist with drafting and refactoring code and polishing the manuscript for readability. The authors reviewed all AI-assisted content. All scientific claims, experimental design, and results reported in the paper were generated and validated by the authors

\bibliographystyle{ACM-Reference-Format}
\bibliography{reference}

@String(ICCV  = {ICCV})

@String(AAAI = {AAAI})

@article{li2025jit,
  title={Back to Basics: Let Denoising Generative Models Denoise},
  author={Li, Tianhong and He, Kaiming},
  journal={arXiv preprint arXiv:2511.13720},
  year={2025}
}

@inproceedings{
loshchilov2018decoupled,
title={Decoupled Weight Decay Regularization},
author={Ilya Loshchilov and Frank Hutter},
booktitle={International Conference on Learning Representations},
year={2019}
}

@article{he2020stnet,
  title={Integrating spatial gene expression and breast tumour morphology via deep learning},
  author={He, Bryan and Bergenstr{\aa}hle, Ludvig and Stenbeck, Linnea and Abid, Abubakar and Andersson, Alma and Borg, {\AA}ke and Maaskola, Jonas and Lundeberg, Joakim and Zou, James},
  journal={Nature biomedical engineering},
  volume={4},
  number={8},
  pages={827--834},
  year={2020},
  publisher={Nature Publishing Group UK London}
}

@inproceedings{
zhu2025stem,
title={Diffusion Generative Modeling for Spatially Resolved Gene Expression Inference from Histology Images},
author={Sichen Zhu and Yuchen Zhu and Molei Tao and Peng Qiu},
booktitle={The Thirteenth International Conference on Learning Representations},
year={2025}
}

@inproceedings{yang2023egn,
  title={Exemplar guided deep neural network for spatial transcriptomics analysis of gene expression prediction},
  author={Yang, Yan and Hossain, Md Zakir and Stone, Eric A and Rahman, Shafin},
  booktitle={Proceedings of the IEEE/CVF Winter Conference on Applications of Computer Vision},
  pages={5039--5048},
  year={2023}
}

@article{xie2023bleep,
  title={Spatially resolved gene expression prediction from histology images via bi-modal contrastive learning},
  author={Xie, Ronald and Pang, Kuan and Chung, Sai and Perciani, Catia and MacParland, Sonya and Wang, Bo and Bader, Gary},
  journal={Advances in Neural Information Processing Systems},
  volume={36},
  pages={70626--70637},
  year={2023}
}

@inproceedings{chung2024triplex,
  title={Accurate spatial gene expression prediction by integrating multi-resolution features},
  author={Chung, Youngmin and Ha, Ji Hun and Im, Kyeong Chan and Lee, Joo Sang},
  booktitle={Proceedings of the IEEE/CVF Conference on Computer Vision and Pattern Recognition},
  pages={11591--11600},
  year={2024}
}

@inproceedings{huang2025stflow,
  title={Scalable Generation of Spatial Transcriptomics from Histology Images via Whole-Slide Flow Matching},
  author={Huang, Tinglin and Liu, Tianyu and Babadi, Mehrtash and Jin, Wengong and Ying, Zhitao},
  booktitle={International Conference on Machine Learning},
  pages={25550--25565},
  year={2025},
  organization={PMLR}
}

@inproceedings{jaume2024hest,
    author = {Guillaume Jaume and Paul Doucet and Andrew H. Song and Ming Y. Lu and Cristina Almagro-Perez and Sophia J. Wagner and Anurag J. Vaidya and Richard J. Chen and Drew F. K. Williamson and Ahrong Kim and Faisal Mahmood},
    title = {HEST-1k: A Dataset for Spatial Transcriptomics and Histology Image Analysis},
    booktitle = {Advances in Neural Information Processing Systems},
    year = {2024},
    month = dec,
}

@article{andersson2021spatial,
  title={Spatial deconvolution of HER2-positive breast cancer delineates tumor-associated cell type interactions},
  author={Andersson, Alma and Larsson, Ludvig and Stenbeck, Linnea and Salm{\'e}n, Fredrik and Ehinger, Anna and Wu, Sunny Z and Al-Eryani, Ghamdan and Roden, Daniel and Swarbrick, Alex and Borg, {\AA}ke and others},
  journal={Nature communications},
  volume={12},
  number={1},
  pages={6012},
  year={2021},
  publisher={Nature Publishing Group UK London}
}

@article{erickson2022spatially,
  title={Spatially resolved clonal copy number alterations in benign and malignant tissue},
  author={Erickson, Andrew and He, Mengxiao and Berglund, Emelie and Marklund, Maja and Mirzazadeh, Reza and Schultz, Niklas and Kvastad, Linda and Andersson, Alma and Bergenstr{\aa}hle, Ludvig and Bergenstr{\aa}hle, Joseph and others},
  journal={Nature},
  volume={608},
  number={7922},
  pages={360--367},
  year={2022},
  publisher={Nature Publishing Group UK London}
}

@article{lake2023atlas,
  title={An atlas of healthy and injured cell states and niches in the human kidney},
  author={Lake, Blue B and Menon, Rajasree and Winfree, Seth and Hu, Qiwen and Melo Ferreira, Ricardo and Kalhor, Kian and Barwinska, Daria and Otto, Edgar A and Ferkowicz, Michael and Diep, Dinh and others},
  journal={Nature},
  volume={619},
  number={7970},
  pages={585--594},
  year={2023},
  publisher={Nature Publishing Group UK London}
}

@article{vicari2024spatial,
  title={Spatial multimodal analysis of transcriptomes and metabolomes in tissues},
  author={Vicari, Marco and Mirzazadeh, Reza and Nilsson, Anna and Shariatgorji, Reza and Bj{\"a}rterot, Patrik and Larsson, Ludvig and Lee, Hower and Nilsson, Mats and Foyer, Julia and Ekvall, Markus and others},
  journal={Nature Biotechnology},
  volume={42},
  number={7},
  pages={1046--1050},
  year={2024},
  publisher={Nature Publishing Group US New York}
}

@misc{arcinstitute2025vcc,
  title = {Virtual Cell Challenge 2025 Wrap-Up},
  author = {{ARC Institute}},
  year = {2025},
  note = {Accessed: 2026-03-30}
}

@article{zeng2022Hist2ST,
  title={Spatial transcriptomics prediction from histology jointly through transformer and graph neural networks},
  author={Zeng, Yuansong and Wei, Zhuoyi and Yu, Weijiang and Yin, Rui and Yuan, Yuchen and Li, Bingling and Tang, Zhonghui and Lu, Yutong and Yang, Yuedong},
  journal={Briefings in Bioinformatics},
  volume={23},
  number={5},
  pages={bbac297},
  year={2022},
  publisher={Oxford University Press}
}

@article{pang2021histogene,
  title={Leveraging information in spatial transcriptomics to predict super-resolution gene expression from histology images in tumors},
  author={Pang, Minxing and Su, Kenong and Li, Mingyao},
  journal={BioRxiv},
  pages={2021--11},
  year={2021},
  publisher={Cold Spring Harbor Laboratory}
}

@article{schmauch2025miso,
  title={A deep learning-based multiscale integration of spatial omics with tumor morphology},
  author={Schmauch, Beno{\^\i}t and Herpin, Lo{\"\i}c and Olivier, Antoine and Duboudin, Thomas and Dubois, R{\'e}my and Gillet, Lucie and Filiot, Alexandre and Schiratti, Jean-Baptiste and Di Proietto, Valentina and Le Corre, Delphine and others},
  journal={Nature Communications},
  year={2025},
  publisher={Nature Publishing Group UK London}
}

@article{liu2025SpatialEx,
  title={High-parameter spatial multi-omics through histology-anchored integration},
  author={Liu, Yonghao and Wang, Chuyao and Wang, Zhikang and Chen, Liang and Li, Zhi and Song, Jiangning and Zou, Qi and Gao, Rui and Qian, Bin-Zhi and Feng, Xiaoyue and others},
  journal={Nature Methods},
  pages={1--14},
  year={2025},
  publisher={Nature Publishing Group US New York}
}

@article{schroeder2025iscale,
  title={Scaling up spatial transcriptomics for large-sized tissues: uncovering cellular-level tissue architecture beyond conventional platforms with iSCALE},
  author={Schroeder, Amelia and Loth, Melanie L and Luo, Chunyu and Yao, Sicong and Yan, Hanying and Zhang, Daiwei and Piya, Sarbottam and Plowey, Edward and Hu, Wenxing and Clemenceau, Jean R and others},
  journal={Nature methods},
  volume={22},
  number={9},
  pages={1911--1922},
  year={2025},
  publisher={Nature Publishing Group US New York}
}

@article{fu2025ghist,
  title={Spatial gene expression at single-cell resolution from histology using deep learning with GHIST},
  author={Fu, Xiaohang and Cao, Yue and Bian, Beilei and Wang, Chuhan and Graham, Dinny and Pathmanathan, Nirmala and Patrick, Ellis and Kim, Jinman and Yang, Jean Yee Hwa},
  journal={Nature methods},
  volume={22},
  number={9},
  pages={1900--1910},
  year={2025},
  publisher={Nature Publishing Group US New York}
}

@inproceedings{
jing2024alphafold,
title={AlphaFold Meets Flow Matching for Generating Protein Ensembles},
author={Bowen Jing and Bonnie Berger and Tommi Jaakkola},
booktitle={Forty-first International Conference on Machine Learning},
year={2024}
}

@inproceedings{zhang2025flowpolicy,
    title={FlowPolicy: Enabling Fast and Robust 3D Flow-Based Policy via Consistency Flow Matching for Robot Manipulation}, 
    volume={39}, 
    booktitle={Proceedings of the AAAI Conference on Artificial Intelligence}, 
    author={Zhang, Qinglun and Liu, Zhen and Fan, Haoqiang and Liu, Guanghui and Zeng, Bing and Liu, Shuaicheng}, 
    year={2025}, 
    month={Apr.}, 
    pages={14754-14762} 
}

@inproceedings{wang2025vaflow,
  title={VAFlow: Video-to-Audio Generation with Cross-Modality Flow Matching},
  author={Wang, Xihua and Cheng, Xin and Wang, Yuyue and Song, Ruihua and Wang, Yunfeng},
  booktitle={Proceedings of the IEEE/CVF International Conference on Computer Vision},
  pages={11777--11786},
  year={2025}
}

@inproceedings{
    jin2025pyramidal,
    title={Pyramidal Flow Matching for Efficient Video Generative Modeling},
    author={Yang Jin and Zhicheng Sun and Ningyuan Li and Kun Xu and Kun Xu and Hao Jiang and Nan Zhuang and Quzhe Huang and Yang Song and Yadong MU and Zhouchen Lin},
    booktitle={The Thirteenth International Conference on Learning Representations},
    year={2025}
}

@article{lu2026pmf,
  title={One-step Latent-free Image Generation with Pixel Mean Flows},
  author={Lu, Yiyang and Lu, Susie and Sun, Qiao and Zhao, Hanhong and Jiang, Zhicheng and Wang, Xianbang and Li, Tianhong and Geng, Zhengyang and He, Kaiming},
  journal={arXiv preprint arXiv:2601.22158},
  year={2026}
}

@inproceedings{
    yu2026scdfm,
    title={sc{DFM}: Distributional Flow Matching Model for Robust Single-Cell Perturbation Prediction},
    author={Chenglei Yu and Chuanrui Wang and Bangyan Liao and Tailin Wu},
    booktitle={The Fourteenth International Conference on Learning Representations},
    year={2026}
}

@article{song2025l1,
  title={L1 Sample Flow for Efficient Visuomotor Learning},
  author={Song, Weixi and Chen, Zhetao and Xu, Tao and Zeng, Xianchao and Zhou, Xinyu and Yang, Lixin and Wang, Donglin and Lu, Cewu and Li, Yong-Lu},
  journal={arXiv preprint arXiv:2511.17898},
  year={2025}
}

@article{zhang2019rmsnorm,
  title={Root mean square layer normalization},
  author={Zhang, Biao and Sennrich, Rico},
  journal={Advances in neural information processing systems},
  volume={32},
  year={2019}
}

@misc{zhang2024relu2,
      title={ReLU$^2$ Wins: Discovering Efficient Activation Functions for Sparse LLMs}, 
      author={Zhengyan Zhang and Yixin Song and Guanghui Yu and Xu Han and Yankai Lin and Chaojun Xiao and Chenyang Song and Zhiyuan Liu and Zeyu Mi and Maosong Sun},
      year={2024},
      eprint={2402.03804},
      archivePrefix={arXiv},
      primaryClass={cs.LG}
}

@article{huang2023visual,
  title={A visual--language foundation model for pathology image analysis using medical twitter},
  author={Huang, Zhi and Bianchi, Federico and Yuksekgonul, Mert and Montine, Thomas J and Zou, James},
  journal={Nature medicine},
  volume={29},
  number={9},
  pages={2307--2316},
  year={2023},
  publisher={Nature Publishing Group US New York}
}

@article{chen2024towards,
  title={Towards a general-purpose foundation model for computational pathology},
  author={Chen, Richard J and Ding, Tong and Lu, Ming Y and Williamson, Drew FK and Jaume, Guillaume and Song, Andrew H and Chen, Bowen and Zhang, Andrew and Shao, Daniel and Shaban, Muhammad and others},
  journal={Nature Medicine},
  volume={30},
  number={3},
  pages={850--862},
  year={2024},
  publisher={Nature Publishing Group US New York}
}

@inproceedings{dit,
  title     = {Scalable Diffusion Models with Transformers},
  author    = {William Peebles and Saining Xie},
  booktitle = ICCV,
  year      = {2023}
}

@article{staahl2016visualization,
  title={Visualization and analysis of gene expression in tissue sections by spatial transcriptomics},
  author={St{\aa}hl, Patrik L and Salm{\'e}n, Fredrik and Vickovic, Sanja and Lundmark, Anna and Navarro, Jos{\'e} Fern{\'a}ndez and Magnusson, Jens and Giacomello, Stefania and Asp, Michaela and Westholm, Jakub O and Huss, Mikael and others},
  journal={Science},
  volume={353},
  number={6294},
  pages={78--82},
  year={2016},
  publisher={American Association for the Advancement of Science}
}

@article{chen2020spatial,
  title={Spatial transcriptomics and in situ sequencing to study Alzheimer’s disease},
  author={Chen, Wei-Ting and Lu, Ashley and Craessaerts, Katleen and Pavie, Benjamin and Frigerio, Carlo Sala and Corthout, Nikky and Qian, Xiaoyan and Lal{\'a}kov{\'a}, Jana and K{\"u}hnemund, Malte and Voytyuk, Iryna and others},
  journal={Cell},
  volume={182},
  number={4},
  pages={976--991},
  year={2020},
  publisher={Elsevier}
}

@article{baccin2020combined,
  title={Combined single-cell and spatial transcriptomics reveal the molecular, cellular and spatial bone marrow niche organization},
  author={Baccin, Chiara and Al-Sabah, Jude and Velten, Lars and Helbling, Patrick M and Gr{\"u}nschl{\"a}ger, Florian and Hern{\'a}ndez-Malmierca, Pablo and Nombela-Arrieta, C{\'e}sar and Steinmetz, Lars M and Trumpp, Andreas and Haas, Simon},
  journal={Nature cell biology},
  volume={22},
  number={1},
  pages={38--48},
  year={2020},
  publisher={Nature Publishing Group UK London}
}

@article{janesick2023high,
  title={High resolution mapping of the tumor microenvironment using integrated single-cell, spatial and in situ analysis},
  author={Janesick, Amanda and Shelansky, Robert and Gottscho, Andrew D and Wagner, Florian and Williams, Stephen R and Rouault, Morgane and Beliakoff, Ghezal and Morrison, Carolyn A and Oliveira, Michelli F and Sicherman, Jordan T and others},
  journal={Nature communications},
  volume={14},
  number={1},
  pages={8353},
  year={2023},
  publisher={Nature Publishing Group UK London}
}

@article{chen2025omiclip,
  title={A visual--omics foundation model to bridge histopathology with spatial transcriptomics},
  author={Chen, Weiqing and Zhang, Pengzhi and Tran, Tu N and Xiao, Yiwei and Li, Shengyu and Shah, Vrutant V and Cheng, Hao and Brannan, Kristopher W and Youker, Keith and Lai, Li and others},
  journal={Nature Methods},
  volume={22},
  number={7},
  pages={1568--1582},
  year={2025},
  publisher={Nature Publishing Group US New York}
}

@article{wang2024m2ort,
  title={M2ORT: many-to-one regression transformer for spatial transcriptomics prediction from histopathology images},
  author={Wang, Hongyi and Du, Xiuju and Liu, Jing and Ouyang, Shuyi and Chen, Yen-Wei and Lin, Lanfen},
  journal={arXiv preprint arXiv:2401.10608},
  year={2024}
}

@article{chadoutaud2026scellst,
  title={sCellST predicts single-cell gene expression from H\& E images},
  author={Chadoutaud, Lo{\"\i}c and Lerousseau, Marvin and Herrero-Saboya, Daniel and Ostermaier, Julian and Fontugne, Jacqueline and Barillot, Emmanuel and Walter, Thomas},
  journal={Nature Communications},
  volume={17},
  number={1},
  pages={1194},
  year={2026},
  publisher={Nature Publishing Group}
}

@inproceedings{
han2026stamp,
title={Fusing Pixels and Genes: Spatially-Aware Learning in Computational Pathology},
author={Minghao Han and Dingkang Yang and Linhao Qu and Zizhi Chen and Gang Li and Han Wang and Jiacong Wang and Lihua Zhang},
booktitle={The Fourteenth International Conference on Learning Representations},
year={2026},
}

@article{jiang2026stwave,
  title = {STWave: Fine‐Scale Spatial Structure Discovery in Microscopic‐Resolution Spatial Transcriptomics via Patchwise Wavelet Graphs},
  author = {Jiang, Tao and Zhang, Songming and Chen, Xiaofeng and Xiao, Xiongtao and Cao, Wenming and Li, Weikai and Zhao, Tong and Li, Bing and Xu, Xinyue and Li, Zhongshan},
  journal = {Advanced Science},
  year = {2026},
  month = aug,
  pages = {e77240},
  publisher = {Wiley},
  doi = {10.1002/advs.77240},
  pmid = {42615345},
}

\appendix
\section{Method Details}
\label{app:method}

\paragraph{\textbf{Hyperparameter Tuning Details}}

We also conduct comprehensive hyperparameter tuning for our GeneFlow:
the dropout rate is chosen from $\{0, 0.2, 0.5\}$, the number of attention heads is searched in $\{1, 3, 6, 9\}$, the DiT depth is set to $\{3, 6, 9, 12\}$, and the sampling steps are selected from $\{1, 2, 5, 10\}$. We further evaluate the influence of different prior distributions, including ZINB, Gaussian, and Zero distribution, together with multiple parameter configurations of the ZINB distribution. All models adopt an early stopping strategy, where training is halted if no performance improvement is observed for 25 epochs.
Early stopping is determined solely by the convergence of the training objective, while the held-out slide is not used during training or early stopping.

\section{Datasets}

\label{app:datasets}
\begin{table}[htb]
    \centering
    \caption{Dataset statistics of spatially resolved transcriptomics dataset.}
    \label{tab:dataset}
    \resizebox{.9\linewidth}{!}{%
    \begin{tabular}{lcccl}
    \toprule
    Oncotree    & Num. of Slides & Tech.   & Slide ID      & Test Slide \\ \midrule
    PRAD        & 23             & Visium  & MEND139-162   & MEND145    \\
    Kidney      & 23             & Visium  & NCBI692-715   & NCBI697    \\
    Mouse Brain & 14             & Visium  & NCBI658-672   & NCBI667    \\
    HER2ST      & 37             & Visium  & SPA119-155    & SPA148     \\ \bottomrule
    \end{tabular}
    }
\end{table}

We evaluated our method on four ST datasets from the HEST-1k database~\citep{jaume2024hest}, covering multiple organs, species, and disease conditions.
All datasets were generated using spot-based ST platforms and provide paired histology images and spatially resolved gene expression profiles.
For each dataset, we adopted a slide-level split, where one slide was held out as the test set and all remaining slides were used for training. The detailed statistics of each dataset are summarized in \cref{tab:dataset}.

\textbf{HER2ST dataset}~\citep{andersson2021spatial} consists of breast cancer tissue sections with spatial spots of 100~$\mu$m diameter.
The dataset contains multiple histopathology slides with a total of 13,594 spatial spots, each associated with gene expression measurements.
The tissue regions include both normal breast tissue and malignant areas.
We used the SPA148 slide as the test set and trained the model on all remaining slides.

\textbf{Kidney Visium dataset}~\citep{lake2023atlas} contains 23 human kidney tissue slides covering three pathological conditions: healthy controls, chronic kidney disease, and acute kidney injury.
The data were generated using Visium technology with a spot diameter of 55~$\mu$m, and each slide contains between 315 and 4,159 spatial spots.
The samples span both cortical and medullary anatomical regions.
We used the NCBI697 slide as the test set and trained on the remaining slides.

\textbf{Healthy Mouse Brain dataset}~\citep{vicari2024spatial} consists of 14 Visium slides from healthy adult mouse brain tissue, with spot sizes of 55~$\mu$m and between 2,675 and 3,617 spots per slide.
This dataset provides spatial gene expression across multiple brain structures.
We selected the NCBI667 slide as the test slide and used the rest of the slides for training.

\textbf{Human Prostate Cancer (PRAD) Visium dataset}~\citep{erickson2022spatially} includes 23 prostate cancer tissue slides sequenced using the 10x Genomics Visium platform, with a spatial resolution of 55~$\mu$m.
The number of spots per slide ranges from 1,418 to 4,079.
We selected the MEND145 slide as the held-out test slide and used the other slides for training.

\section{Baselines}
\label{app:baselines}

We compare our method with several state-of-the-art approaches for spatial gene expression prediction, including StNet~\citep{he2020stnet}, EGN~\citep{yang2023egn}, BLEEP~\citep{xie2023bleep}, TRIPLEX~\citep{chung2024triplex}, Stem~\citep{zhu2025stem}, and STFlow~\citep{huang2025stflow}.

\begin{itemize}[leftmargin=*]
    \item
    StNet adopts a convolutional neural network as the image encoder to extract visual features from spot images.
    It maps the learned visual features to gene expression profiles through a simple linear projection head.

    \item
    EGN introduces an exemplar-guided network structure for accurate ST analysis.
    It achieves effective prediction by learning from representative and informative samples.

    \item
    BLEEP learns joint representations of histology images and gene expressions via contrastive learning.
    It predicts gene expressions by retrieving and averaging expressions of similar reference spots.

    \item
    TRIPLEX designs a multi-branch encoder architecture to capture target, local, and global information.
    It fuses multi-scale spatial features to achieve comprehensive transcriptomics prediction from histology images.

    \item
    Stem integrates histology image features and gene embeddings as conditions within a DiT-based generative architecture.
    It generates accurate spatial gene expressions by conditioning on image patches through a transformer-based diffusion model.

    \item
    STFlow constructs a flow-based generative framework for ST prediction.
    It leverages conditional flow matching to model the distribution of gene expressions from histology images.
\end{itemize}

\begin{figure}[tb]
    \centering
    \includegraphics[width=.95\linewidth]{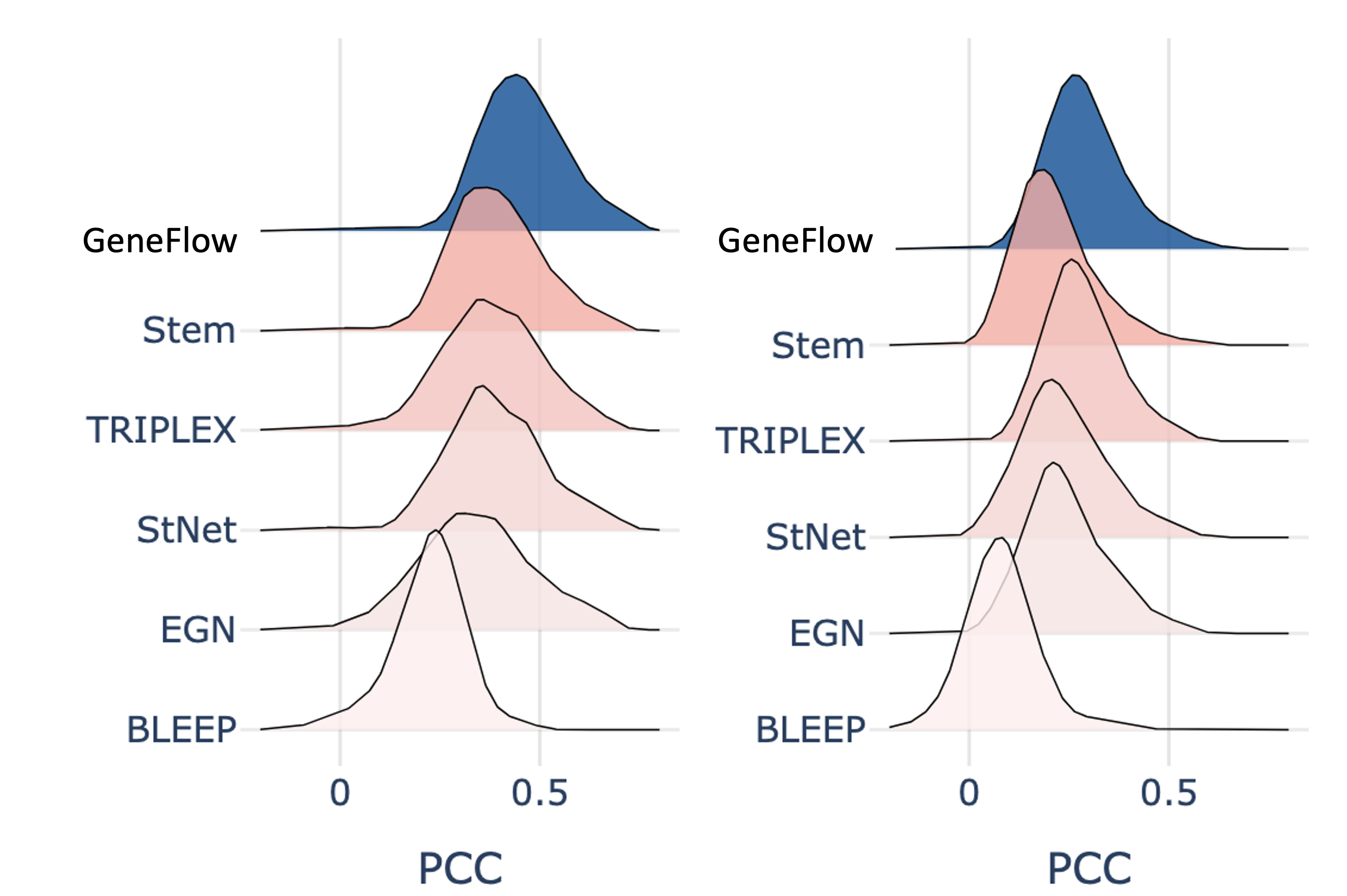}
    \caption{The distribution of PCC scores across all predicted genes for each model on MEND145 and NCBI667.}
    \label{fig:corr_ridge}
\end{figure}

\begin{table}[tb]
    \centering
    \caption{Performance metrics under different DiT depths.}
    \label{tab:dit_depth}
    \begin{tabular}{c|cccc}
    \toprule
    DiT depth & PCC-10 & PCC-50 & PCC-200 & PCC-1000\\
    \midrule
    3  & 0.5761 & 0.5109 & 0.4192 & 0.2616 \\
    6  & 0.5716 & 0.5122 & 0.4208 & 0.2545 \\
    9  & 0.5743 & 0.5180 & 0.4243 & 0.2633 \\
    12 & 0.5743 & 0.5121 & 0.4252 & 0.2739\\
    \bottomrule
    \end{tabular}
 \end{table}

\begin{figure*}[tb]
    \centering
    \includegraphics[width=.95\linewidth]{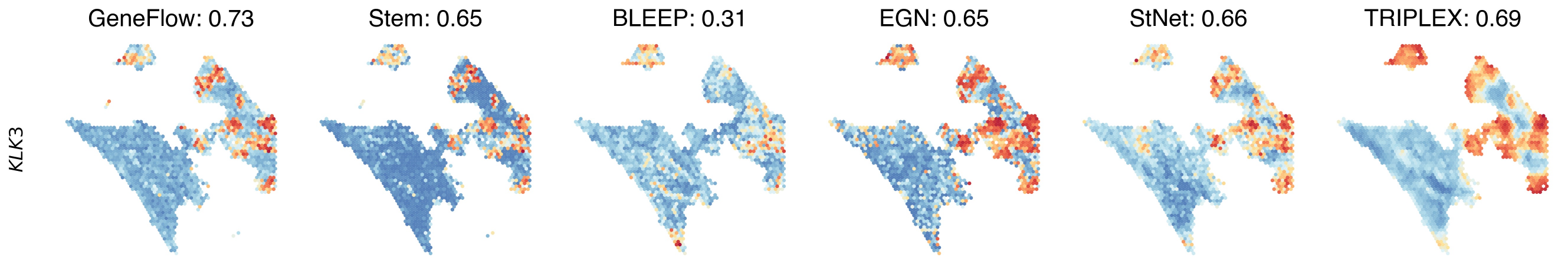}
    \includegraphics[width=.95\linewidth]{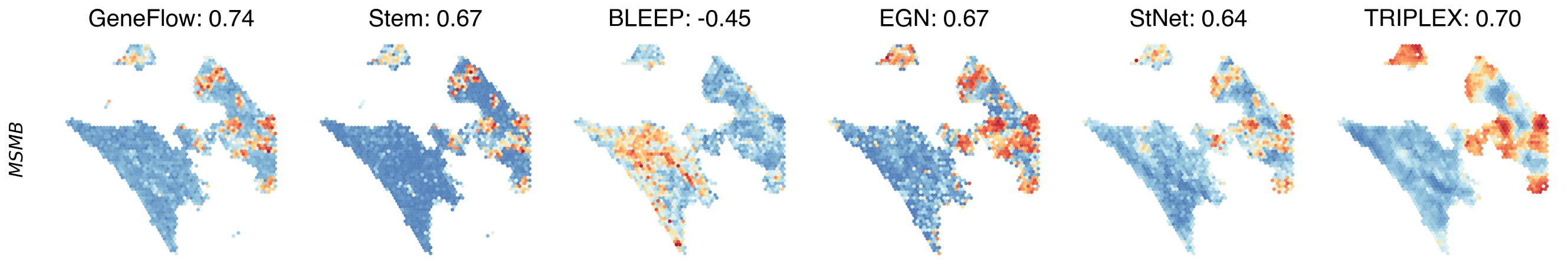}
    \caption{Visualization of biomarker genes.
    Predicted expression of \textit{KLK3} and \textit{MSMB} in PRAD tissue slide MEND145.
    }
    \label{fig:biomarker_2}
\end{figure*}

\section{Additional Experimental Analysis}
\label{app:analysis}

\paragraph{\textbf{Ridge Plot Analysis}}

As shown in \cref{fig:corr_ridge}, GeneFlow maintains a clear lead in overall gene expression prediction performance: its distribution (blue curve) is significantly right-shifted compared to all competing models, with the highest density peak concentrated in the high-PCC range (centered above 0.5). This indicates that GeneFlow achieves stronger correlation between predicted and ground-truth expression values for the majority of genes, with performance tightly clustered in the optimal range. In contrast, the distributions of Stem, TRIPLEX, StNet, EGN, and BLEEP are progressively left-shifted, with lower peak positions and wider spreads, reflecting weaker predictive accuracy and greater variability across genes. Notably, BLEEP and EGN exhibit extended tails toward lower PCC values, further highlighting their inferior global performance relative to GeneFlow. Collectively, these patterns confirm GeneFlow's superior and consistent ability to capture the global distribution of spatial gene expression.

\paragraph{\textbf{Effectiveness of Gene Tokenizer}}

\begin{figure}[H]
    \centering
    \includegraphics[width=.95\linewidth]{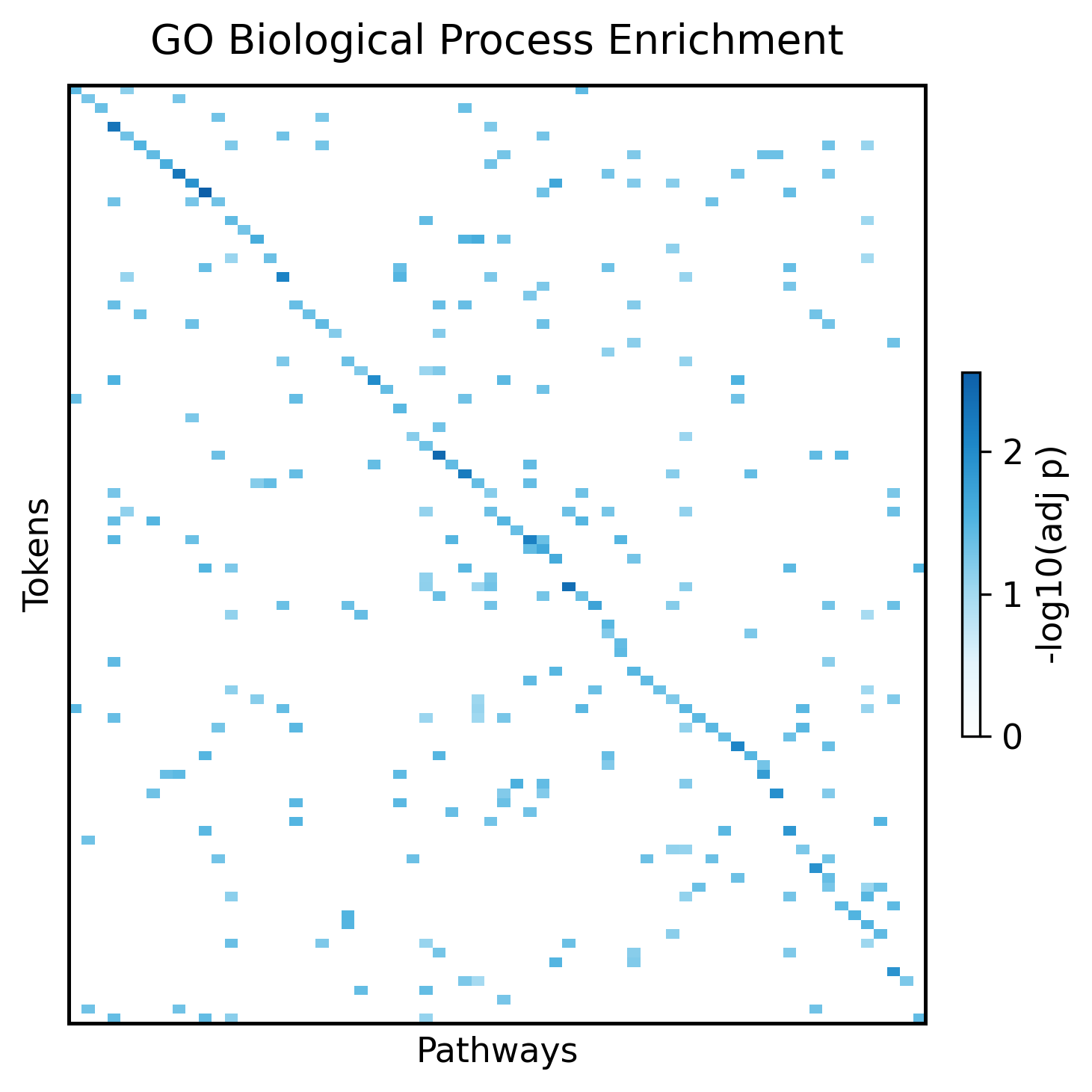}
    \caption{GO Biological Process Enrichment heatmap for learned gene tokens in GeneFlow.}
    \label{fig:enrichment}
\end{figure}

We perform Gene Ontology (GO) Biological Process enrichment analysis to examine the functional structure captured by the learned gene module tokens. For each token $j$, we calculate the importance score of gene $g$ as the $\ell_2$ norm of its corresponding projection weights across the token dimension, i.e., $s_{g,j}=\|W_{g,j}\|_2$, and select the five genes with the highest scores. Enrichment analysis is conducted using the \texttt{GSEApy} Python package with its default enrichment configuration, mouse-specific GO Biological Process annotations, and default background gene universe. We use the multiple-testing-adjusted $p$-values returned by \texttt{GSEApy}, considering terms with an adjusted $p$-value below $0.05$ to be statistically significant. The heatmap in \cref{fig:enrichment} reports the corresponding $-\log_{10}$-transformed adjusted $p$-values.

The enrichment patterns suggest that the gene tokenizer learns biologically meaningful and functionally distinct gene representations. Each token is associated with a largely non-overlapping set of enriched GO Biological Process terms, indicating that different tokens capture relatively coherent functional gene modules. In addition, the enrichment matrix is highly sparse, with most \texttt{(token, pathway)} pairs not reaching statistical significance. This suggests that individual tokens tend to focus on specific biological functions rather than indiscriminately mixing multiple processes. Overall, these results provide evidence that the gene tokenizer captures structured and interpretable gene programs suitable for ST modeling.

\begin{table}[tb]
\centering
\caption{Performance under different hidden dimensions of gene tokenizer}
\label{tab:hidden_dim}
    \begin{tabular}{crcccc}
    \toprule
    Dim. $H$  & PCC-10  & PCC-50  & PCC-200 & PCC-1000    \\
    \midrule
    12     & 0.5502  & 0.4872  & 0.3919  & 0.2299   \\
    24      & 0.5888  & 0.5277  & 0.4424  & 0.2874  \\
    36      & 0.5851  & 0.5254  & 0.4369  & 0.2834  \\
    48     & 0.5856  & 0.5238  & 0.4287  & 0.2723 \\
    96      & 0.5741  & 0.5123  & 0.4237  & 0.2725   \\
    \bottomrule
    \end{tabular}
    \vspace{-3mm}
\end{table}

\paragraph{\textbf{Effect of DiT Depth}}
\cref{tab:dit_depth} shows the effect of DiT depth on performance. Increasing the depth generally improves the PCC metrics, particularly PCC-200 and PCC-1000, indicating that deeper architectures better capture long-range dependencies across genes. Although increasing the depth from 9 to 12 provides only marginal gains and causes slight fluctuations in PCC-10 and PCC-50, it achieves the best performance on the larger gene panels. Therefore, we select a depth of 12 to prioritize large-panel prediction performance despite its limited marginal benefit.

\paragraph{\textbf{Effect of ZINB Distribution Hyperparameters}}

We use a zero-inflated negative binomial distribution $\mathcal{Z}(\mu,\phi,\pi)$ to model the sparsity and overdispersion of ST expression counts. During training, the initial expression $\x_0$ is sampled directly from $\mathcal{Z}(\mu,\phi,\pi)$ and interpolated with the ground-truth expression $\x_1$ as $\x_t=(1-t)\x_0+t\x_1$. During inference, a sample drawn from the same prior serves as the initial expression estimate and is iteratively refined by GeneFlow.

We fix the zero-inflation probability at $\pi=0.5$, since an excessively large value would make the prior approach a near-zero distribution. We vary the mean $\mu\in\{0.1,0.2,0.4\}$ and dispersion $\phi\in\{1,2,4\}$ to evaluate the sensitivity of GeneFlow to the ZINB parameterization. As shown in \cref{tab:prior_zinb}, the results are relatively stable for PCC-10 and PCC-50, while larger gene panels exhibit greater sensitivity to the prior parameters. Overall, a smaller mean combined with moderate dispersion provides better performance, with $(\mu,\phi)=(0.1,1)$ achieving the best PCC-1000 score.

\subsection{Loss Space Analysis}
\label{app:loss_space}

\begin{figure}[htb]
    \centering
    \includegraphics[width=.95\linewidth]{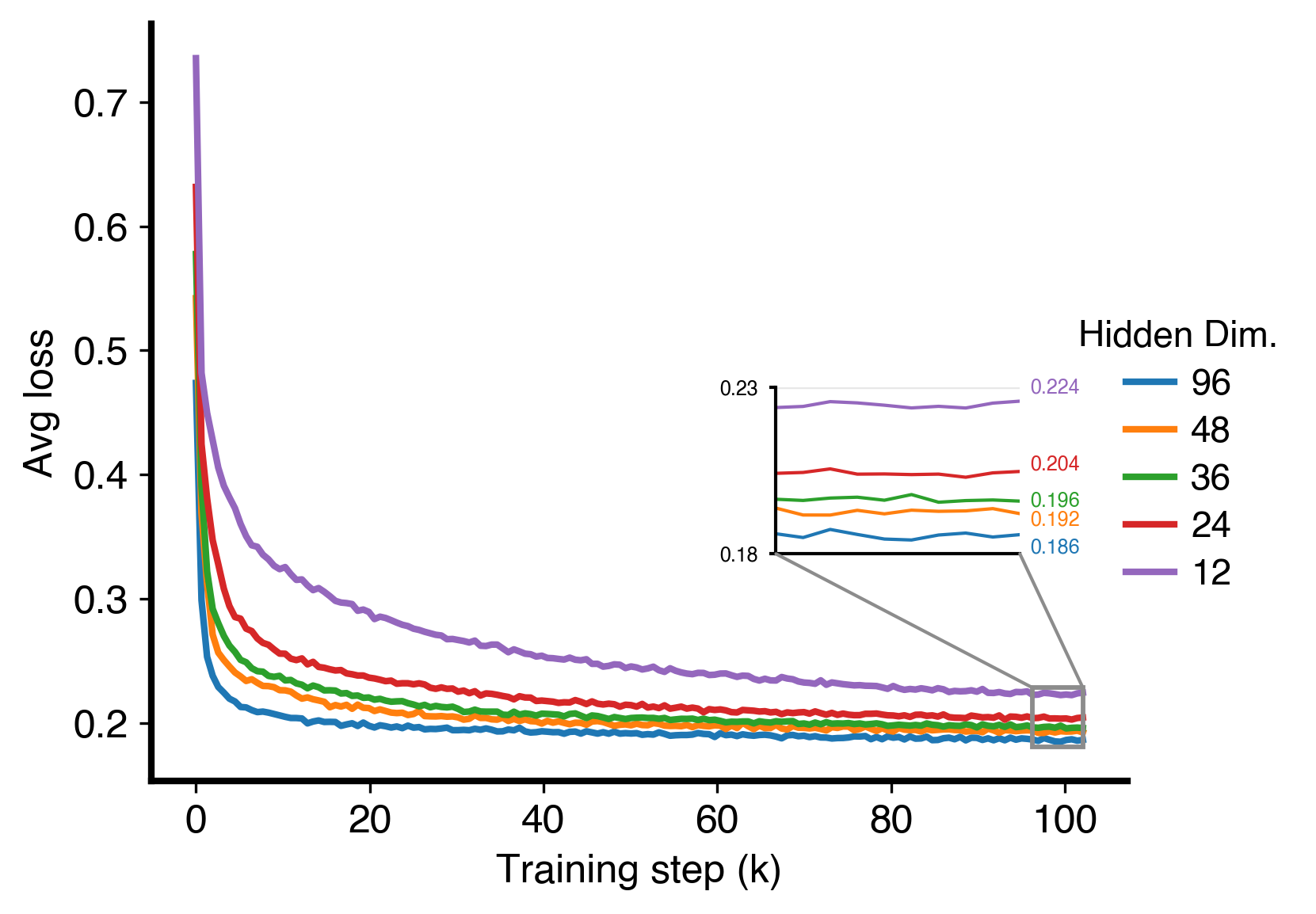}
    \caption{Training loss curves across different hidden dimensions of the gene tokenizer.}
    \label{fig:loss_over_steps}
\end{figure}

As shown in \cref{fig:loss_over_steps}, increasing the hidden dimension monotonically reduces the converged training loss, indicating that larger models can better fit the FM objective in the velocity space. However, the best predictive accuracy is achieved at a moderate dimension (24), while further increasing capacity degrades performance, suggesting that minimizing $\mathcal{L}_{\text{FM}}$ is not well aligned with the downstream metric and cannot reliably reflect generalization. It indicates the need for more effective training-time evaluation protocols that are better correlated with downstream prediction quality, rather than relying solely on the FM loss as a proxy.

\end{document}